\documentclass{article}
\PassOptionsToPackage{numbers}{natbib}

 \usepackage[main, final]{neurips_2026}

\usepackage[utf8]{inputenc} %
\usepackage[T1]{fontenc}    %
\usepackage{hyperref}       %
\usepackage{url}            %
\usepackage{booktabs}       %
\usepackage{amsfonts}       %
\usepackage{nicefrac}       %
\usepackage{microtype}      %
\usepackage{xcolor}         %
\usepackage{colortbl}        %
\usepackage{graphicx}
\usepackage{multirow}
\usepackage{multicol}
\usepackage{bm}
\usepackage{dsfont} %
\newcommand{\mathbbm}[1]{\mathds{#1}}
\usepackage{algorithm} %
\usepackage{amssymb}
\usepackage{amsmath}
\usepackage{cleveref}
\crefname{section}{Sec.}{Secs.}                         
\crefname{figure}{Fig.}{Figs.}                            
\crefname{table}{Tab.}{Tabs.}                           
\crefname{equation}{Eq.}{Eqs.}    
\usepackage{pifont}
\usepackage{wrapfig}
\definecolor{VividRed}{HTML}{FF6B6B}
\definecolor{VividGreen}{HTML}{2ED573}
\def\mtd{CoPhy}

\def\mone{\textcolor{black}{cognitive prior distillation}}
\def\mtwo{\textcolor{black}{physical world model rollout}}
\def\mthree{\textcolor{black}{cognitive-physical policy optimization}}

\title{Distill to Think, Foresee to Act: Cognitive-Physical Reinforcement Learning for Autonomous Driving}

\author{Yang Wu$^{1}$ \quad Qiang Meng$^{\dagger}$ \quad Zhaojiang Liu$^3$ \quad Youquan Liu$^4$ \quad Jian Yang$^{1,2}$ \quad Jin Xie$^{2,*}$ \\[0.2cm]
$^1$PCA Lab, Nanjing University of Science and Technology \\
$^2$Nanjing University \quad $^3$Shanghai Jiao Tong University \quad $^4$Fudan University \\
\texttt{wuyang98@njust.edu.cn} \quad \texttt{csjxie@nju.edu.cn}
}

\begin{document}

\maketitle
\begingroup
\renewcommand{\thefootnote}{\ensuremath{\dagger}}
\footnotetext[0]{Independent researcher. $^{*}$ Corresponding author.}
\endgroup

\begin{abstract}
  Current end-to-end autonomous driving models are fundamentally constrained by the behavioral cloning ceiling of imitation learning. While reinforcement learning offers a path to smarter autonomy, it demands two missing pieces of infrastructure: (1) a cognitive foundation that understands traffic semantics and driving intent, and (2) a foresighted physical environment that can anticipate the consequences of candidate actions.
  To this end, we propose \textbf{\mtd{}}, a \textbf{Co}gnitive-\textbf{Phy}sical reinforcement learning framework for autonomous driving.
  To \emph{distill to think}, we distill VLM knowledge into the BEV encoder and then discard the VLM entirely, so that at inference time the VLM is not required while the plug-in interface is kept for language command inputs.
  To \emph{foresee to act}, we build an auto-regressive BEV world model that explicitly predicts future semantic maps conditioned on candidate actions, serving as an interpretable physical sandbox from which safety metrics are directly derived.
  Building upon this dual infrastructure, we optimize the driving policy via GRPO with a novel dual-reward mechanism: a \emph{physical} reward derived from BEV rollouts enforces hard safety constraints, while a \emph{cognitive} reward from a language-aligned scorer ensures intent compliance.
  Extensive experiments demonstrate that \mtd{} not only achieves state-of-the-art results on NAVSIM v1 and v2 benchmarks, but also enables safer driving via cognitively informed scene compliance and flexible intent control through user-defined language instructions.
  Project page: \url{https://wuyang98.github.io/CoPhy/}.

  %
  %
  %
  %
  %
  %

\end{abstract}

\vspace{-1.2em}
\section{Introduction}
\begin{figure*}[!t]
\centering
\includegraphics[width=0.98\textwidth]{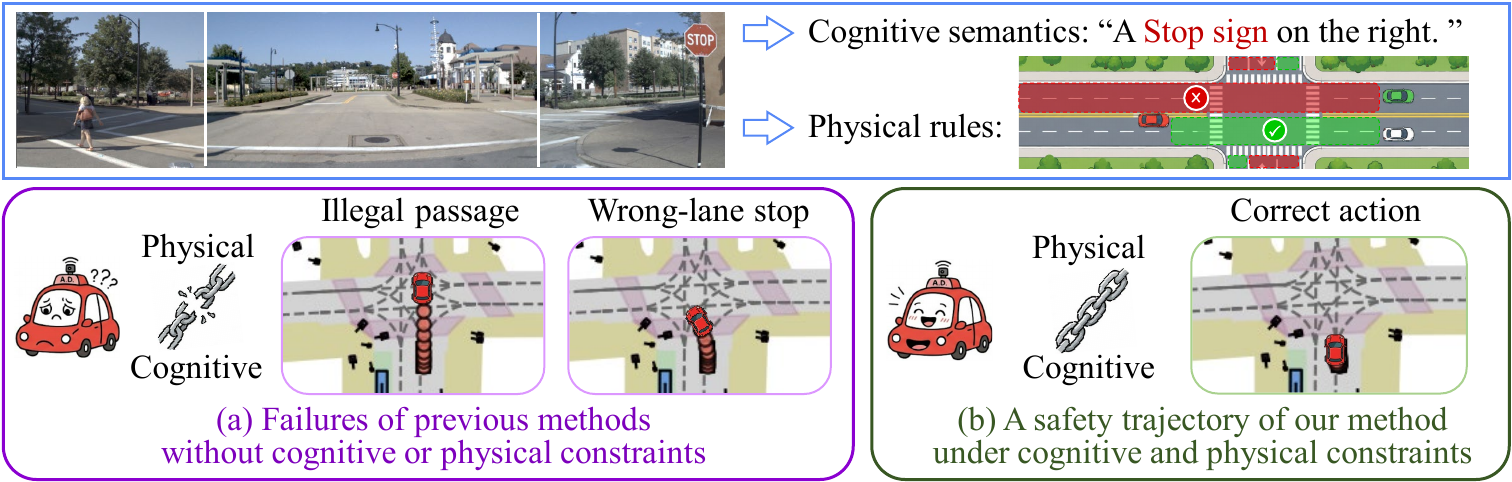}
\vspace{-0.8em}
\caption{Safe driving requires jointly reasoning about \emph{cognitive} cues, such as traffic rules and driver intent, and \emph{physical} constraints, such as spatial feasibility and collision risk. Neglecting the former leads to semantic violations (e.g., ignoring a STOP sign), while neglecting the latter leads to spatial violations (e.g., stopping in the oncoming lane). \textbf{(a)} Previous methods isolate the two and therefore frequently produce such failures. \textbf{(b)} Our method enforces both constraints jointly, so that the output trajectory follows the cognitive semantics and obeys reasonable physical rules.}
\label{fig: intro}
\vspace{-2.0em}
\end{figure*}

\vspace{-0.5em}
\label{sec: intro}
End-to-end (E2E) autonomous driving (AD) has advanced rapidly in recent years~\cite{chitta2022transfuser, hu2023planning, chen2024end}, learning driving policies directly from raw sensor inputs.
Yet the dominant training paradigm, behavior cloning on expert demonstrations, introduces a fundamental bottleneck: models are largely confined to fitting predefined rules and expert trajectories~\cite{chen2024end, zhao2025survey}, severely restricting their safety and adaptability in open-world scenarios.
A key limitation is the lack of cognitive scene understanding.
Existing E2E methods~\cite{chitta2022transfuser, hu2023planning, jiang2023vad, li2025end, liao2025ddrive} rely on pure visual features, skewing optimization towards short-sighted trajectory fitting and leaving them unable to comprehend complex traffic scenarios or incorporate driver intentions.
Vision-Language Models (VLMs)~\cite{park2024vlaad, renz2025simlingo, jiang2024senna, zhou2025autovla, luo2025adathinkdrive} offer a promising pathway for scene reasoning.
However, their prohibitive computational cost hinders edge deployment. 
Moreover, current VLM-based approaches over-rely on textual capabilities, either reducing driving to question answering or naively attaching an action head to the language model. Consequently, they overlook safety issues at the execution level, leaving semantic intent and physical execution misaligned, a limitation further compounded by reasoning hallucinations.
As highlighted in \cref{fig: intro}, isolating cognitive semantics from physical rules inevitably leads to critical driving failures, such as ignoring traffic signals or violating spatial constraints.
This calls for a lightweight mechanism that equips the E2E model with rich cognitive priors without online VLM overhead.

Yet cognitive understanding alone is insufficient for safe driving.
Regardless of how scene semantics are acquired, existing paradigms lack explicit physical-world reasoning to anticipate risks, such as collisions and lane violations.
Similar to human drivers who mentally simulate outcomes before acting, an autonomous system should be able to imagine the consequences of candidate actions and proactively avoid dangerous trajectories.
While some methods~\cite{li2025end, yan2025ad, yang2025raw2drive, rowe2025poutine} utilize implicit world models or simulators~\cite{dauner2024navsim} to predict future latent features, they fundamentally detach from interpretable physical states, failing to provide actionable safety assessment.
This motivates an explicit world model that rolls out future BEV states as a ``what-if'' physical sandbox for trajectory evaluation.

Furthermore, even with cognitive and physical foresight in place, models trained purely via imitation learning remain trapped within the behavioral cloning paradigm: the learned policy can only replicate the behaviors it has observed, without the ability to discover alternative strategies that may be equally or more effective~\cite{chen2024end}.
This limitation manifests as distributional brittleness: when confronted with out-of-distribution scenarios, the policy lacks the exploratory experience needed to generalize.
Reinforcement learning (RL) offers a principled mechanism to overcome this ceiling by allowing the agent to explore beyond expert data and optimize its policy through trial-and-error interaction.
However, applying RL to E2E driving poses distinct challenges: it requires both a faithful environment for rollouts and meaningful reward signals.
While recent approaches~\cite{jiang2025alphadrive, li2026drive, zhou2025autovla} have begun exploring RL for driving, a unified paradigm that integrates soft cognitive preferences (e.g., intent compliance) with hard physical constraints (e.g., collision avoidance) remains absent.

In this work, we propose \textbf{\mtd{}}, a cognitive-physical learning framework that addresses all the above challenges in a unified architecture.
To bridge the cognitive gap, we introduce a \mone{} that distills rich offline language priors into BEV representations, endowing the E2E model with scene understanding at zero additional inference cost.
To enable physical foresight, we establish a \mtwo{} that explicitly simulates future spatiotemporal evolutions conditioned on candidate actions, serving as a physical sandbox for ``what-if'' trajectory rollouts and safety assessment.
To overcome the limitations of imitation learning, we propose \mthree{}. Driven by Group Relative Policy Optimization (GRPO), it explores the policy space via a dual-reward mechanism that synergizes hard physical constraints derived from BEV rollouts with soft cognitive preferences from a distilled language-aligned scorer.
By harmonizing these components, \textbf{\mtd{}} achieves a driving policy that is cognitively aware, physically safe, and intention-aligned.
Extensive experiments on NAVSIM v1 and v2 benchmarks demonstrate that \textbf{\mtd{}} sets a new state-of-the-art, achieving 91.4 PDMS and 86.1 EPDMS respectively, with leading safety-critical metrics across all sub-scores. Moreover, the distilled cognitive channel naturally enables intent-controllable driving via user language commands.
Our main contributions are summarized as:
\begin{itemize}
    \item We propose \textbf{\mtd{}}, a cognitive-physical framework for E2E AD. It grounds cognitive priors into spatial perception at zero inference cost via a \mone{}, while establishing an explicit ``what-if'' sandbox through a \mtwo{}.
    \item To fully exploit this predictive sandbox and break the traditional imitation ceiling, we introduce a \mthree{} that leverages a GRPO-driven dual-reward mechanism to balance hard physical constraints with soft cognitive preferences.
    \item Extensive experiments on NAVSIM v1 and v2 demonstrate that \textbf{\mtd{}} achieves state-of-the-art performance with 91.4 PDMS and 86.1 EPDMS, alongside leading safety-critical sub-scores and an inherent intent-controllable driving capability.
\end{itemize}


\vspace{-1.1em}
\section{Related Works}
\vspace{-0.2em}
\label{sec: relate}

\noindent\textbf{End-to-end Autonomous Driving.}
End-to-end autonomous driving advances rapidly by integrating multi-task perception and planning into unified architectures~\cite{prakash2021multi, hu2023planning, jiang2023vad}. Recent innovations span diverse algorithmic paradigms. For instance, generative and diffusion-based models~\cite{zheng2024genad, liao2025ddrive, li2025recogdrive, xu2025wam} significantly enhance trajectory robustness, while world models~\cite{li2025enhancing, li2025end, xia2026drivelaw, zhang2025epona, wu2026gem} model future scene evolution to support driving prediction and planning. Complementary sensor-level generation methods synthesize diverse weather conditions in LiDAR point clouds~\cite{wu2025weathergen}. Concurrently, efforts to optimize computational efficiency introduce state-space models like Mamba~\cite{yuan2024drama} and multi-teacher knowledge distillation~\cite{li2024hydra}. Nevertheless, driving policies trained through behavioral cloning remain fundamentally bottlenecked by the inherent brittleness and poor interpretability characteristic of pure imitation learning.

\noindent\textbf{VLMs for Autonomous Driving.}
Vision-Language Models significantly elevate autonomous driving via robust semantic understanding, predominantly following two technical routes. The first emphasizes semantic reasoning, leveraging visual question answering~\cite{jin2023adapt, park2024vlaad, renz2025simlingo, sima2024drivelm, jiang2024senna} and chain-of-thought processes~\cite{wang2025alpamayo, jiang2025alphadrive, liu2025x} for high-level decision-making. The second paradigm targets direct trajectory generation by appending action heads to foundational models~\cite{hwang2024emma, zhou2026opendrivevla, shao2024lmdrive, li2025drivevla}, unifying diverse tasks within a single linguistic space. Beyond policy learning, language also enables controllable scene synthesis, such as text-guided LiDAR generation~\cite{wu2024text2lidar}. However, these language-centric driving architectures largely overlook the interplay between cognitive semantics and rigid physical constraints. While recent efforts~\cite{li2025spacedrive, li2026unidrivevla} infuse spatial awareness via positional encoding, they critically lack an explicit physical rollout mechanism to verify safety consequences.

\noindent\textbf{Reinforcement Learning in Autonomous Driving.}
Following its success in foundational language models~\cite{guo2025deepseek} and robotic systems~\cite{tang2025deep}, reinforcement learning aligns autonomous driving policies. 
%
%
DriveDPO~\cite{shang2025drivedpo} explicitly leverages it to unearth robust strategies, while broader vision-language-action frameworks~\cite{jiang2025alphadrive, rowe2025poutine, li2026drive} similarly bridge semantic reasoning with strategic planning.
For closed-loop exploration, predictive approaches~\cite{sheng2026explorevla, yang2025raw2drive} derive intrinsic rewards from scene imaginations, while language-guided frameworks~\cite{zhou2025autovla, jiang2025irl} refine low-level execution using semantic feedback. 
Crucially, these paradigms isolate physical constraints from cognitive preferences, necessitating a unified dual-reward formulation to jointly optimize deterministic safety and intent compliance.


\vspace{-0.5em}
\section{Methodology}
\begin{figure*}[!t]
\includegraphics[width=1.0\textwidth]{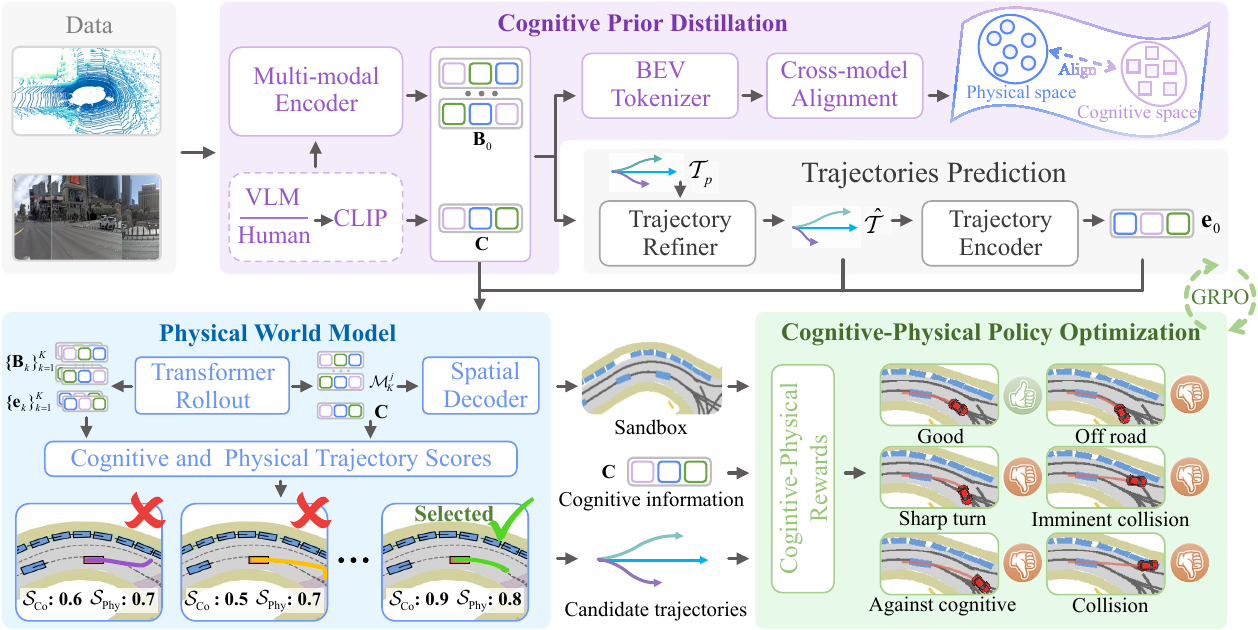}
\vspace{-1.3em}
\caption{Overview of \mtd{}. 
Multi-modal data are encoded into BEV state $\textbf{B}_0$. 
\textbf{(a)} The \mone{} module aligns $\textbf{B}_0$ with VLM-derived cognitive feature $\textbf{C}$ via cross-modal distillation, and injects $\textbf{C}$ into the trajectory refiner to produce cognitively-aware candidates. 
\textbf{(b)} The \mtwo{} module auto-regressively rolls out future BEV states over $K$ steps as a physical sandbox; trajectories are then scored along cognitive and physical axes via hierarchical selection. 
\textbf{(c)} The \mthree{} optimizes the policy with dual cognitive-physical rewards.
}
\label{fig: flowchart}
\vspace{-1.1em}
\end{figure*}

\vspace{-0.5em}
\label{sec: method}
The core idea of \mtd{} is to build a cognitive-physical learning framework, jointly ensuring physical safety and cognitive scene understanding for autonomous driving. 
As illustrated in \cref{fig: flowchart}, it comprises three modules:
(1) \mone{} that injects language priors into the end-to-end model,
(2) \mtwo{} that enables spatiotemporal reasoning of driving scenes,
and (3) \mthree{} that breaks the imitation learning ceiling via dual rewards. 
We detail the training in \cref{sec: vlm,sec: rl} and the inference pipeline in \cref{sec: pipeline}.

\vspace{-1.0em}
\subsection{Cognitive-Physical Imitation Learning}
\vspace{-0.5em}
\label{sec: vlm}

Our end-to-end model takes multi-modal sensor inputs $\mathcal{V} = \{\mathcal{V}_{\text{RGB}}, \mathcal{V}_{\text{LiDAR}}\}$ and projects them via a multi-modal encoder $\mathcal{E}_{\bm{\theta}}$ to obtain the BEV state $\textbf{B}_0 = \mathcal{E}_{\bm{\theta}}(\mathcal{V})$. 
Separately, $N$ kinematic trajectory prototypes $\mathcal{T}_{p}$ are pre-extracted by clustering expert demonstrations~\cite{chen2024vadv2, chitta2022transfuser, li2025end}, constraining the action space within physically feasible bounds.
A trajectory refiner $\pi_\theta$ then yields candidate trajectories $\hat{\mathcal{T}} = \mathcal{T}_{p} + \pi_\theta(\texttt{TE}(\mathcal{T}_{p}), \textbf{B}_{0})$, where $\texttt{TE}$ is the trajectory encoder. 
We will later augment this formulation with cognitive conditioning in \cref{eq:traj}.
The candidates are supervised by the expert trajectory $\mathcal{T}$ via
\begin{equation}
    \mathcal{L}_{\text{Traj}} = \|\hat{\mathcal{T}} - \mathcal{T}\|_1.
\end{equation}

\noindent\textbf{Distill to Think: Cognitive Prior Distillation.}
Vision-Language Models (VLMs) excel at high-order reasoning, benefiting cognitive understanding in current VLA methods.
However, their prohibitive computational cost hinders edge deployment in autonomous driving systems.
Therefore, we distill the VLM's cognitive ability into the end-to-end model during training and discard the VLM entirely at inference.
Specifically, we fine-tune InternVL3-8B~\cite{zhu2025internvl3} via LoRA~\cite{hu2022lora} across diverse driving VQA datasets~\cite{malla2023drama, qian2024nuscenes, xu2024drivegpt4, sima2024drivelm, marcu2024lingoqa}, making it a domain-expert reasoning engine optimized to output cognitive priors such as traffic signs, road topologies, and driver intentions. 
Conditioned on visual inputs $\mathcal{V}$ and textual prompt $Q$, this domain-specific cognition is supervised by
\vspace{-0.5em}
\begin{equation}
    \mathcal{L}_{\texttt{VLM}} = -\sum_{l=1}^{L} \log P_{\bm{\xi}}(y_l \mid y_{<l}, \mathcal{V}, Q),
\end{equation}
\vspace{-0.5em}
where $y_l$ is the $l$-th token of the target sequence and $\bm{\xi}$ denotes the LoRA parameters.

Once trained, the VLM is frozen and used to generate textual descriptions during IL training, which are encoded by a CLIP text encoder~\cite{radford2021learning} into the cognitive prior $\textbf{C}$.
To bridge the cognitive-physical gap, we force $\mathcal{E}_{\bm{\theta}}$ to internalize these language priors via a cross-modal cosine distance objective:
\begin{equation}
    \label{eq:distill}
    \mathcal{L}_{\texttt{Distill}} = 1 - \texttt{BE}(\mathbf{B}_0)\cdot \textbf{C}/(\|\texttt{BE}(\mathbf{B}_0)\|_2 \|\textbf{C}\|_2),
\end{equation}
where $\texttt{BE}$ is a BEV tokenizer to extract the cognitive feature from the BEV state $\mathbf{B}_0$.

In addition to the feature-level distillation, we further inject $\textbf{C}$ into the trajectory refiner by
\begin{equation}
    \label{eq:traj}
  \hat{\mathcal{T}} = \mathcal{T}_{p} + \pi_\theta(\texttt{TE}(\mathcal{T}_{p}) , \textbf{B}_{0}, \mathbf{C}).
\end{equation}
This not only enriches trajectories with cognitive awareness, but also opens an interface for intent-controllable driving via language commands, an underexplored capability of great practical value.

\noindent\textbf{Foresee to Act: Physical World Model Rollout.}
While cognitive distillation enriches the model with semantic understanding, it alone cannot address the core safety challenges of autonomous driving. 
Equally critical is the ability to imagine future states and anticipate potential dangers, enabling the model to proactively avoid unsafe trajectories.
Moreover, breaking the imitation learning ceiling requires closed-loop exploration, where the agent reasons about consequences of diverse actions, a core principle underlying reinforcement learning.
Thus, we introduce an auto-regressive world model that serves two purposes: (1) providing a physical sandbox to assess candidate trajectories via future rollouts, and (2) acting as the environment for reinforcement learning-based policy optimization.

Unlike conventional methods constrained by external simulators or purely latent features~\cite{rowe2025poutine, li2025end, yang2025raw2drive}, our world model operates directly within the BEV space, auto-regressively rolling out futures conditioned on both spatial geometries and cognitive priors.
Specifically, a Transformer-based~\cite{vaswani2017attention} world model takes the current BEV state $\textbf{B}_0$, the cognitive feature $\textbf{C}$, and $N$ initial ego features $\textbf{e}_{0} = \texttt{TE}(\hat{\mathcal{T}})$ as input. 
To predict future states and actions, the model recursively unrolls the transitions:
\begin{equation}
    \label{eq:wm}
    (\textbf{B}_{k}, \textbf{e}_{k}) = \texttt{WM}( \textbf{B}_{k-1}, \textbf{e}_{k-1}, \textbf{C} ),
\end{equation}
where $k \in \{1, \cdots, K\}$ denotes the future time step.
Since optimizing all future states is computationally expensive during training, we randomly sample an index $j \sim \mathcal{U}(1, N)$ in each iteration and use a lightweight spatial decoder to project the corresponding state $\textbf{B}_{K}^{j}$ into dense BEV semantic maps $\mathcal{\hat{M}}_{K}^{j} \in \mathbb{R}^{H \times W \times C_{c}}$, where $C_{c}$ denotes the category count. 
Using nuPlan~\cite{caesar2021nuplan} to generate ground-truth maps $\mathcal{M}^{j}_{K}$~\cite{li2025end}, we train the physical world model by supervising the predictions via Focal Loss~\cite{lin2017focal}:
\begin{equation}
    \mathcal{L}_{\texttt{WM}} = \texttt{FocalLoss}(\mathcal{\hat{M}}_{K}^{j}, \mathcal{M}^{j}_{K}).
\end{equation}

\begin{figure*}[!t]
\includegraphics[width=1.0\textwidth]{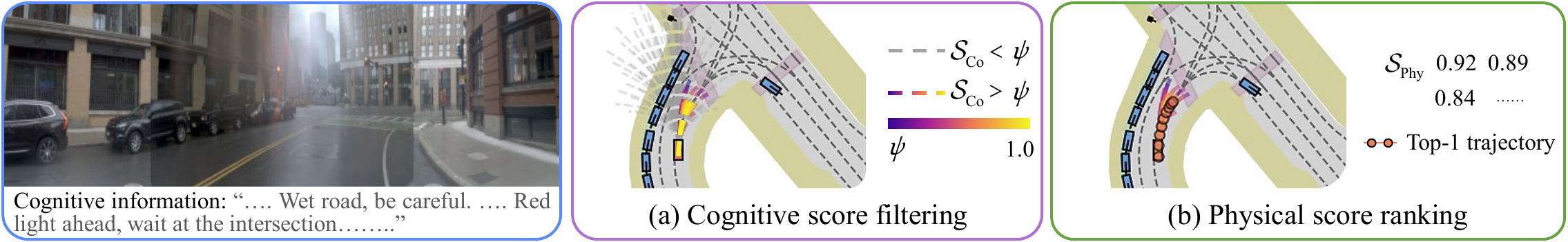}
\vspace{-1.5em}
\caption{Hierarchical trajectory selection. \textbf{(a)} Candidates passing the cognitive threshold $\hat{s}_{\texttt{Co}} \ge \psi$ are retained and subsequently \textbf{(b)} ranked by the physical score $\hat{s}_{\texttt{Phy}}$ to determine the optimal trajectory.}
\label{fig: traj_select}
\vspace{-1.3em}
\end{figure*}

\noindent\textbf{Cognitive and Physical Trajectory Scores.}
With trajectory generation and cognitive-physical representations in place, a key question remains: how to score the $N$ predicted trajectories.
We propose to score trajectories along two complementary axes, physical safety and cognitive compliance, and then select the optimal one.

The first criterion is the physical score~\cite{li2025end}, which relates to the distance $d$ to the expert trajectory, yielding $s_{\texttt{Phy}} \propto \exp(-d)$.
To predict this score, we concatenate the BEV states $\{\textbf{B}_{k}\}_{k=1}^{K}$ along the temporal dimension and compact them into a BEV vector $\textbf{b}$ via a few convolutional layers.
The vector, along with the $K$-step ego features $\{\textbf{e}_{k}\}_{k=1}^{K}$, is projected into the physical score $\hat{s}_{\texttt{Phy}} \in \mathbb{R}^{B \times N}$ via MLP and softmax.
This explicitly favors expert-aligned behaviors and is optimized via cross-entropy:
\begin{equation}
    \label{eq:phy_loss}
    \hat{s}_{\texttt{Phy}} = \texttt{Softmax}\big(\texttt{MLP}([\textbf{b};\; \textbf{e}])\big),\quad \mathcal{L}_{\texttt{Phy}} = -s_{\texttt{Phy}} \log (\hat{s}_{\texttt{Phy}}).
\end{equation}

\vspace{-0.8em}
The second criterion is the cognitive alignment between $\textbf{C}$ and the candidate trajectory $\hat{\mathcal{T}}$, defined as the cosine similarity after projecting $\textbf{C}$ and $\hat{\mathcal{T}}$ through two MLP layers.
Since precise human labels are unavailable, we assume that the expert trajectory aligns well with the VLM's cognitive intent, and supervise $\hat{s}_{\texttt{Co}}\in \mathbb{R}^{B \times N}$ via a contrastive InfoNCE loss~\cite{radford2021learning, oord2018representation}.
\begin{equation}
    \label{eq:co_loss}
    \hat{s}_{\texttt{Co}} = \texttt{cos}\big( \texttt{MLP}(\hat{\mathcal{T}}),\; \texttt{MLP}(\textbf{C}) \big),\quad \mathcal{L}_{\texttt{Co}} = -\log \frac{\exp(s^{+} / \epsilon)} {\exp(s^{+} / \epsilon) + \sum_{m=1}^{5} \exp(s^{-}_{m} / \epsilon)},
\end{equation}
where $s^{+}$ is the similarity coupling the expert trajectory with $\textbf{C}$, $s^{-}_{m}$ are the negative similarities from the 5 candidates with largest $L_2$ distances to the expert, and $\epsilon$ is a temperature hyperparameter.

\noindent\textbf{Imitation Learning.}
Both \cref{eq:phy_loss,eq:co_loss} encourage the highest-scoring trajectory to match the expert, forming the imitation learning objective in essence.
Together with the trajectory, distillation, and world model losses, all modules except the VLM are jointly trained in an end-to-end manner via
\begin{equation}
\label{eq: loss}
    \mathcal{L}_{\texttt{IL}} = \mathcal{L}_{\texttt{Traj}} + \mathcal{L}_{\texttt{Distill}} + \mathcal{L}_{\texttt{WM}} +  \mathcal{L}_{\texttt{Phy}} +  \mathcal{L}_{\texttt{Co}}.
\end{equation}

\vspace{-1.0em}
\subsection{Cognitive-Physical Policy Optimization}
\label{sec: rl}
\vspace{-0.5em}

While imitation learning (IL) provides a strong initialization, it implicitly assumes that the expert trajectory is optimal in both physical safety and cognitive coherence, an assumption that does not always hold.
Moreover, the model never explores beyond the expert data, limiting its ability to generalize to unseen scenarios or surpass expert-level performance.
Reinforcement learning (RL) offers a natural solution to break this ceiling, yet it typically requires a well-established environment and meaningful rewards, both of which are difficult to build in end-to-end autonomous driving.


Fortunately, after the IL stage, the trained world model can already roll out future BEV states for any candidate trajectory, naturally serving as an internal driving environment.
Building on this foundation, we employ RL to optimize only the trajectory refiner $\pi_\theta$ via \mthree{}, while freezing all other parameters.
This greatly reduces the training cost, and also prevents the environment (embedded in the world model parameters) from co-evolving with the policy, which would lead to training collapse.
We next detail the designed rewards and the training objective.

\noindent\textbf{Physical Reward.}
Though the physical score in \cref{eq:phy_loss} can evaluate the trajectory itself, it derives from the expert trajectory without explicitly measuring physical safety, a property extremely important in autonomous driving.
Therefore, we propose to derive the physical reward $\mathcal{R}_{\texttt{Phy}}$ directly from the world model's imagined BEV semantic maps $\mathcal{\hat{M}}_{k}$. By projecting the ego-vehicle's spatial footprint $\mathcal{F}_{\text{ego}}$ onto these rollouts, we formulate $\mathcal{R}_{\texttt{Phy}}$ as a weighted summation of five core metrics:
\begin{equation}
    \mathcal{R}_{\texttt{Phy}}^{i} = \omega_{1} \mathcal{R}_{\texttt{NC}}^{i} + \omega_{2} \mathcal{R}_{\texttt{DAC}}^{i} + \omega_{3} \mathcal{R}_{\texttt{EP}}^{i} + \omega_{4} \mathcal{R}_{\texttt{TTC}}^{i} + \omega_{5} \mathcal{R}_{\texttt{Comf}}^{i},
\end{equation}
where $i$ indexes the candidate trajectory and $\omega_{1\dots5}$ are balancing coefficients between components: 


\textit{(1) No Collision:} $\mathcal{R}_{\texttt{NC}} = \mathbbm{1}[\mathcal{F}_{\text{ego}} \cap \mathcal{M}_{\text{obs}} = \emptyset]$ strictly avoids predicted obstacle channels $\mathcal{M}_{\text{obs}}$. 

\textit{(2) Drivable Area:} $\mathcal{R}_{\texttt{DAC}} = \mathbbm{1}[\mathcal{F}_{\text{ego}} \subseteq \mathcal{M}_{\text{drive}}]$ strictly confines the ego vehicle within the predicted road boundary channels $\mathcal{M}_{\text{drive}}$ to guarantee spatial safety.

\textit{(3) Ego Progress:} To encourage adaptive efficiency, we compute $\mathcal{R}_{\texttt{EP}} = \sigma\big((x_{K} - \mu_v)/\sigma_v\big)$, where $x_{K}$ is the final longitudinal displacement at the $K$-th step, $\mu_v = v_{\text{ego}} \Delta T$ ($\Delta T=4.0\text{s}$ in our implementation) is the expected progress, and $\sigma_v = \max(0.3 \mu_v, 2.0)$ is the tolerance margin.

\textit{(4) Time-to-Collision:} To promote proactive safety margins, we reward maintaining a safe headway: $\mathcal{R}_{\texttt{TTC}} = \sigma\big((d_{\min} - d_{\text{safe}})/\Delta d\big)$, where $d_{\min}$ is the minimal $L_2$ distance to surrounding agents, with threshold $d_{\text{safe}} = 5.0\text{m}$ and scaling factor $\Delta d = 2.0\text{m}$.

\textit{(5) Comfort:} To ensure smooth kinematic execution, we penalize harsh dynamics via exponential decay: $\mathcal{R}_{\texttt{Comf}} = \exp(-\bar{a}/\lambda_a) \cdot \exp(-\bar{j}/\lambda_j)$, where $\bar{a}$ and $\bar{j}$ are the mean acceleration and jerk magnitudes over the $K$-step horizon, normalized by $\lambda_a=2.0$ and $\lambda_j=5.0$.

\begin{figure*}[!t]
\includegraphics[width=1.0\textwidth]{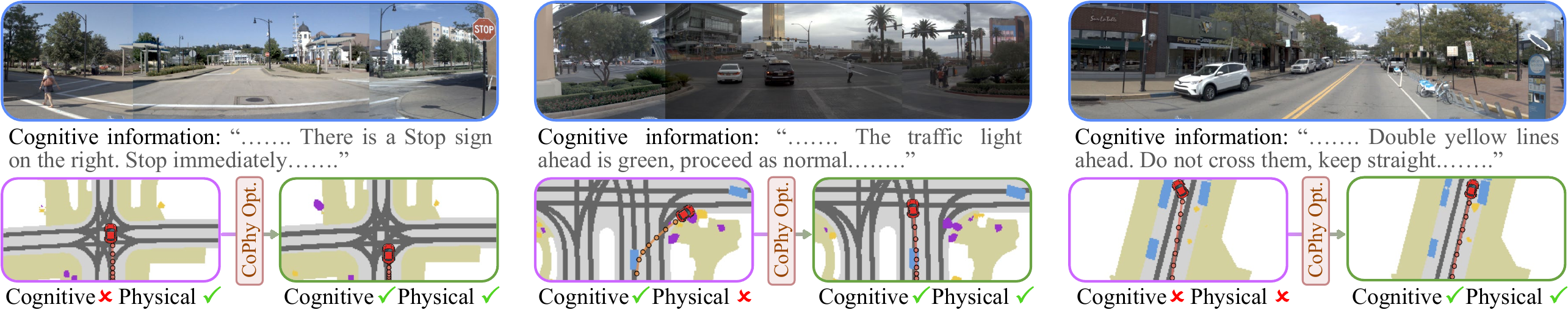}
\vspace{-1.5em}
\caption{Qualitative comparisons of trajectories before and after optimization. The dual-reward mechanism systematically rectifies cognitive lapses, such as ignoring a STOP sign, alongside physical infractions, such as crossing double yellow lines, thereby achieving rigorous joint compliance.}
\label{fig: rl_reward}
\vspace{-1.3em}
\end{figure*}

\noindent\textbf{Cognitive Reward.}
Physical metrics alone cannot capture high-level semantic compliance such as intent adherence. 
We therefore leverage the cognitive scorer trained during the IL phase to provide language-aligned feedback.
For a sampled trajectory $\hat{\mathcal{T}}^i$, the cognitive reward is computed as:
\begin{equation}    
    \mathcal{R}_{\texttt{Co}}^i = \texttt{cos}\big( \texttt{MLP}(\hat{\mathcal{T}}^i),\; \texttt{MLP}(\textbf{C}) \big).
\end{equation}
where the $\texttt{MLP}$ projectors are frozen from the IL-trained language-aligned scorer.
\cref{fig: rl_reward} shows how the dual rewards jointly rectify both cognitive lapses and physical infractions.

\noindent\textbf{Policy Optimization via GRPO.}
We optimize the policy via Group Relative Policy Optimization (GRPO)~\cite{shao2024deepseekmath}, which estimates advantages from intra-group reward statistics without requiring a separate critic network.
At each iteration, the old policy $\pi_{\bm{\theta}_{\text{old}}}$ samples a group of $G$ trajectories $\{\hat{\mathcal{T}}^{1}, \dots, \hat{\mathcal{T}}^{G}\}$. The unified reward for each candidate combines both components: $\mathcal{R}^{i} = \mathcal{R}_{\texttt{Phy}}^{i} + \mathcal{R}_{\texttt{Co}}^{i}$.
GRPO estimates the advantage $\hat{A}^{i}$ via intra-group normalization: $\hat{A}^{i} = (\mathcal{R}^{i} - \mu_{\mathcal{R}}) / (\sigma_{\mathcal{R}} + \varphi)$, where $\mu_{\mathcal{R}}$ and $\sigma_{\mathcal{R}}$ are the group's reward mean and standard deviation, and $\varphi$ ensures numerical stability.
Finally, $\pi_{\bm{\theta}}$ is updated over $E$ epochs by maximizing the clipped surrogate objective. To prevent catastrophic forgetting of expert priors and encourage exploration, we augment the objective with the imitation learning loss $\mathcal{L}_{\texttt{IL}}$ (\cref{eq: loss}) and an entropy bonus $\mathcal{H}(\pi_{\bm{\theta}})$:
\begin{equation}
    \mathcal{J}_{\texttt{GRPO}}(\bm{\theta}) = \frac{1}{G} \sum_{i=1}^{G} \left[ \min \Big( \rho^i \hat{A}^{i}, \texttt{clip}(\rho^i, 1-\gamma, 1+\gamma) \hat{A}^{i} \Big) \right] + \mathcal{H}(\pi_{\bm{\theta}}) - 0.2 \mathcal{L}_{\texttt{IL}},
\end{equation}
where $\rho^i = \pi_{\bm{\theta}}(\hat{\mathcal{T}}^i) / \pi_{\bm{\theta}_{\text{old}}}(\hat{\mathcal{T}}^i)$ is the importance sampling ratio, and the clipping threshold is $\gamma = 0.2$.

\vspace{-0.4em}
\subsection{Inference Pipeline}
\label{sec: pipeline}
\vspace{-0.6em}

We answer three practical questions that arise during inference: (1) how to select the final trajectory from multiple candidates, (2) how to remove the computationally expensive VLM without losing cognitive capability, and (3) how to enable intent-controllable driving via user language commands.

\noindent\textbf{Hierarchical Trajectory Selection.}
For trajectory selection, we propose a hierarchical strategy that scores trajectories along two complementary axes, cognitive compliance and physical safety, rather than a rigid weighted summation.
Specifically, we first filter intention-violating candidates via a cognitive threshold $\psi$ (\cref{fig: traj_select}(a)), forming a valid subset $\Omega = \{i \mid \hat{s}_{\texttt{Co}}^{i} \ge \psi\}$.
The optimal trajectory is then selected as the one with the highest physical score within $\Omega$: $i^* = \mathop{\arg\max}_{i \in \Omega} \hat{s}_{\texttt{Phy}}^{i}$. This ensures intent compliance while independently maximizing physical safety. As shown in \cref{fig: traj_select}(b), this mechanism steers trajectories toward an optimal cognitive-physical equilibrium.

\noindent\textbf{VLM-Free Inference.}
Although the VLM provides powerful scene understanding, deploying it online incurs prohibitive computational cost.
We therefore discard the VLM entirely at inference time, which however means the cognitive feature $\mathbf{C}$ is no longer available.
This would render the trajectory refiner, world model, and cognitive scorer non-functional, as they all require $\mathbf{C}$ as input.
Fortunately, the distillation objective in \cref{eq:distill} encourages $\texttt{BE}(\mathbf{B}_0)$ to align with $\mathbf{C}$, so the BEV encoder already captures the necessary cognitive information.
We thus simply substitute $\texttt{BE}(\mathbf{B}_0)$ for $\mathbf{C}$ in \cref{eq:traj,eq:wm,eq:co_loss} during inference, achieving zero additional overhead.

\noindent\textbf{Intent-Controllable Driving via User Commands.}
In practice, an autonomous system may not always satisfy human preferences, and users may wish to issue specific commands such as ``speed up'', ``change lane'' or ``pull over.''
Our framework naturally supports such intent-controllable driving by replacing the cognitive input source.
Instead of using the distilled $\texttt{BE}(\mathbf{B}_0)$, we encode the user's command via the CLIP text encoder to produce $\mathbf{C}$, which finally achieves intent-controllable driving.

\vspace{-0.8em}
\section{Experiments}
\vspace{-0.5em}
\label{sec: exp}

\subsection{Experimental Setup}
\vspace{-0.5em}
\label{sec: exp_datasets}

\noindent\textbf{Implementation Details.}
Our \mtd{} is trained in three stages.
First, we fine-tune InternVL3-8B~\cite{zhu2025internvl3} with LoRA~\cite{hu2022lora} on a mixture of driving VQA datasets~\cite{malla2023drama, qian2024nuscenes, xu2024drivegpt4, sima2024drivelm, marcu2024lingoqa} for 5 epochs with learning rate $2\text{e-4}$.
Second, with the VLM frozen, we train the world model and policy via imitation learning for 30 epochs with learning rate $5\text{e-5}$.
Finally, with all other modules frozen, we optimize the trajectory refiner via reinforcement learning for 10 epochs with learning rate $1\text{e-5}$.
The reward weights $\omega_{1\dots5}$ are set to $1, 1, 0.4, 0.4, 0.2$ for NC, DAC, EP, TTC, and Comf, respectively, reflecting the relative importance of each driving criterion in NAVSIM~\cite{dauner2024navsim}. 
The group size $G$ is set to 8 in GRPO.
All stages use AdamW~\cite{loshchilov2017decoupled} with linear warmup and cosine annealing~\cite{loshchilov2017sgdr} on 8 H20 GPUs.

\noindent\textbf{Datasets.}
Our method is evaluated on NAVSIM~\cite{dauner2024navsim, caopseudo}, a planning-oriented dataset built upon OpenScene~\cite{openscene2023} with 1,192 training and 136 testing scenarios.
We report results on two simulation versions: v1, which employs non-reactive log replay, and the more challenging v2, which features reactive traffic agents that respond to the ego vehicle's actions in a pseudo-closed-loop setting.

\noindent\textbf{Evaluation Metrics.}
For NAVSIM v1, the primary metric is the Predictive Driver Model Score (PDMS), which aggregates five sub-metrics: No At-fault Collision (NC), Drivable Area Compliance (DAC), Ego Progress (EP), Time-to-Collision (TTC), and Comfort (C). The PDMS is formulated as:
$\text{PDMS} = \text{NC} \times \text{DAC} \times (5 \times \text{EP} + 5 \times \text{TTC} + 2 \times \text{C}) / 12$.
For NAVSIM v2, four additional sub-metrics are introduced: Driving Direction Compliance (DDC), Traffic Lights Compliance (TL), Lane Keeping (LK), and Extended Comfort (EC). The overall score becomes the Extended PDMS (EPDMS):
$\text{EPDMS} = \text{NC} \times \text{DAC} \times \text{DDC} \times \text{TL} \times (5 \times \text{TTC} + 2 \times \text{C} + 5 \times \text{EP} + 2 \times \text{LK} + 2 \times \text{EC}) / 16$.

\begin{table*}[t!]
\centering
\caption{Quantitative comparison with state-of-the-art methods on NAVSIM v1 NAVTEST.
$\texttt{C}$: Camera, $\texttt{L}$: LiDAR. WM: World Model.
Best results are in \textbf{bold}, second best are \underline{underlined}.}
\label{tab: v1}
\resizebox{0.93\textwidth}{!}{
\renewcommand{\arraystretch}{0.93}
\begin{tabular}{l|>{\scriptsize}c|c|c|ccccc|c}
\toprule
Method & \normalsize Venue & Input & WM & NC $\uparrow$ & DAC $\uparrow$ & EP $\uparrow$ & TTC $\uparrow$ & C $\uparrow$ & PDMS $\uparrow$ \\
\midrule
UniAD~\cite{hu2023planning} & CVPR'23 & $\texttt{C}$ & \textcolor{VividRed}{\ding{55}} & 97.8 & 91.9 & 78.8 & 92.9 & \textbf{100} & 83.4 \\
PARA-Drive~\citep{weng2024drive} & CVPR'24 & $\texttt{C}$ & \textcolor{VividRed}{\ding{55}} & 97.9 & 92.4 & 79.3 & 93.0 & 99.8 & 84.0 \\
TransFuser~\citep{prakash2021multi} & CVPR'21 & $\texttt{C}$+$\texttt{L}$ & \textcolor{VividRed}{\ding{55}} & 97.7 & 92.8 & 79.2 & 92.8 & \textbf{100} & 84.0 \\
LAW~\citep{li2025enhancing} & ICLR'25 & $\texttt{C}$ & \textcolor{VividGreen}{\ding{51}} & 96.4 & 95.4 & 81.7 & 88.7 & \underline{99.9} & 84.6 \\
DRAMA~\citep{yuan2024drama} & arXiv'24 & $\texttt{C}$+$\texttt{L}$ & \textcolor{VividRed}{\ding{55}} & 98.0 & 93.1 & 80.1 & 94.8 & \textbf{100} & 85.5 \\
HydraMDP++~\cite{li2025hydra++} & arXiv'25 & $\texttt{C}$ & \textcolor{VividRed}{\ding{55}} & 97.6 & 96.0 & 80.4 & 93.1 & \textbf{100} & 86.6 \\
ARTEMIS~\cite{feng2025artemis} & RAL'25 & $\texttt{C}$+$\texttt{L}$ & \textcolor{VividRed}{\ding{55}} & 98.3 & 95.1 & 81.4 & 94.3 & \textbf{100} & 87.0 \\
World4Drive~\citep{zheng2025world4drive} & ICCV'25 & $\texttt{C}$ & \textcolor{VividGreen}{\ding{51}} & 97.4 & 94.3 & 79.9 & 92.8 & \textbf{100} & 85.1 \\
Epona~\citep{zhang2025epona} & ICCV'25 & $\texttt{C}$ & \textcolor{VividGreen}{\ding{51}} & 97.9 & 95.1 & 80.4 & 93.8 & \underline{99.9} & 86.2 \\
TrajHF~\cite{li2025finetuning} & arXiv'25 & $\texttt{C}$+$\texttt{L}$ & \textcolor{VividGreen}{\ding{51}} & 96.6 & 96.6 & 84.5 & 92.1 & \textbf{100} & 87.6 \\
GoalFlow~\cite{xing2025goalflow} & CVPR'25 & $\texttt{C}$+$\texttt{L}$ & \textcolor{VividRed}{\ding{55}} & 98.3 & 93.8 & 79.8 & 94.3 & \textbf{100} & 85.7 \\
DiffusionDrive~\cite{liao2025ddrive} & CVPR'25 & $\texttt{C}$+$\texttt{L}$ & \textcolor{VividRed}{\ding{55}} & 98.2 & 96.2 & 82.2 & 94.7 & \textbf{100} & 88.1 \\
WoTE~\cite{li2025end} & ICCV'25 & $\texttt{C}$+$\texttt{L}$ & \textcolor{VividGreen}{\ding{51}} & \underline{98.5} & 96.8 & 81.9 & 94.9 & \underline{99.9} & 88.3 \\
ResAD~\cite{zheng2025resad} & CVPR'26 & $\texttt{C}$+$\texttt{L}$ & \textcolor{VividRed}{\ding{55}} & 98.0 & 97.5 & 83.3 & 94.1 & \textbf{100} & 88.8 \\
DriveLaW~\cite{xia2026drivelaw} & CVPR'26 & $\texttt{C}$ & \textcolor{VividGreen}{\ding{51}} & \textbf{99.0} & 97.1 & 81.3 & \underline{96.7} & \textbf{100} & 89.1 \\
VADv2~\cite{chen2024vadv2} & ICLR'26 & $\texttt{C}$ & \textcolor{VividRed}{\ding{55}} & 98.3 & 97.4 & 82.3 & 95.7 & \textbf{100} & 89.3 \\
DriveSuprim~\cite{yao2026drivesuprim} & AAAI'26 & $\texttt{C}$+$\texttt{L}$ & \textcolor{VividRed}{\ding{55}} & 97.8 & 97.3 & \textbf{86.7} & 93.6 & \textbf{100} & 89.9 \\
DriveDPO~\cite{shang2025drivedpo} & NeurIPS'25 & $\texttt{C}$+$\texttt{L}$ & \textcolor{VividRed}{\ding{55}} & \underline{98.5} & \underline{98.1} & 84.3 & 94.8 & \underline{99.9} & \underline{90.0} \\
\midrule
\cellcolor{green!7}\textbf{\mtd{} (Ours)} & \cellcolor{green!7}-- & \cellcolor{green!7}$\texttt{C}$+$\texttt{L}$ & \cellcolor{green!7}\textcolor{VividGreen}{\ding{51}} & \cellcolor{green!7}\textbf{99.0} & \cellcolor{green!7}\textbf{98.2} & \cellcolor{green!7}\underline{85.3} & \cellcolor{green!7}\textbf{96.8} & \cellcolor{green!7}\textbf{100} & \cellcolor{green!7}\textbf{91.4} \\
\bottomrule
\end{tabular}
}
\vspace{-1.5em}
\end{table*}

\begin{table*}[t!]
\centering
\caption{
    Quantitative comparison with state-of-the-art methods on NAVSIM v2 NAVTEST. 
    }
\label{tab: v2}
\resizebox{\textwidth}{!}{
\begin{tabular}{l|>{\scriptsize}c|ccccccccc|c}
\toprule
{Method} & \normalsize Venue & NC $\uparrow$ & DAC $\uparrow$ & DDC $\uparrow$ & TL $\uparrow$ & EP $\uparrow$ & TTC $\uparrow$ & LK $\uparrow$ & C $\uparrow$ & EC $\uparrow$ & EPDMS $\uparrow$ \\
\midrule
Transfuser~\cite{chitta2022transfuser}~\ & CVPR'21 & 96.9 & 89.9 & 97.8 & 99.7 & 87.1 & 95.4 & 92.7 & \underline{98.3} & 87.2 & 76.7 \\
HydraMDP++~\cite{li2025hydra++} & arXiv'25 & 97.2 & \underline{97.5} & \underline{99.4} & 99.6 & 83.1 & 96.5 & 94.4 & 98.2 & 70.9 & 81.4 \\
DriveSuprim~\cite{yao2026drivesuprim} & AAAI'26 & 97.5 & 96.5 & 99.4 & 99.6 & 88.4 & 96.6 & 95.5 & 98.3 & 77.0 & 83.1 \\
ARTEMIS ~\cite{feng2025artemis} & RAL'25 & 98.3 & 95.1 & 98.6 & \underline{99.8} &81.5 & \underline{97.4} & 96.5 & \underline{98.3} & - & 83.1 \\
DiffusionDrive~\cite{liao2025ddrive} & CVPR'25 & 98.2 &95.9 & \underline{99.4} & \underline{99.8} & \underline{87.5} & 97.3 & \underline{96.8} & \underline{98.3} & \underline{87.7} & 84.5\\
ResAD~\cite{zheng2025resad} & CVPR'26 & 97.8 & 97.2 & \textbf{99.5} & \underline{99.8} & \textbf{88.2} & 96.9 & \textbf{97.0} & \textbf{98.4} & \textbf{88.2} & 85.5 \\
\midrule
\cellcolor{green!7}\textbf{\mtd{} (Ours)} & \cellcolor{green!7}-- & \cellcolor{green!7}\textbf{99.3} & \cellcolor{green!7}\textbf{97.6} & \cellcolor{green!7}97.3 & \cellcolor{green!7}\textbf{100} & \cellcolor{green!7}83.6 & \cellcolor{green!7}\textbf{99.1} & \cellcolor{green!7}\underline{96.8} & \cellcolor{green!7}97.4 & \cellcolor{green!7}80.0 & \cellcolor{green!7}\textbf{86.1} \\
\bottomrule
\end{tabular}
}
\vspace{-1.1em}
\end{table*}

\vspace{-0.5em}
\subsection{Main Results}
\vspace{-0.5em}
\label{sec: exp_plan}
In this part, we compare \mtd{} against 18 recent methods spanning diverse paradigms, including pure imitation learning, reinforcement learning, and world model-based approaches.
Both quantitative results and qualitative visualizations are provided to demonstrate the effectiveness of our framework.


\noindent\textbf{Quantitative Comparison.} \cref{tab: v1} benchmarks \mtd{} against state-of-the-art methods on NAVSIM v1. \mtd{} establishes a new state-of-the-art with a PDMS of \textbf{91.4}, surpassing the second-best DriveDPO by \textbf{+1.4}. Crucially, it leads all safety-critical sub-metrics (NC 99.0, DAC 98.2, TTC 96.8). While \mtd{} (85.3) trails the pure IL method DriveSuprim (86.7) in EP, this baseline exhibits pronounced safety degradations (NC 97.8, TTC 93.6). This discrepancy exposes a fundamental IL limitation: aggressively fitting expert progress without explicit safety constraints. Conversely, our \mthree{}, grounded in BEV world model rollouts, delivers robust safety guarantees fundamentally unattainable by pure imitation.

We further evaluate on the more challenging NAVSIM v2 and report results in \cref{tab: v2}. \mtd{} achieves the highest EPDMS of \textbf{86.1}. Safety-critical metrics show even larger gains than v1: NC reaches 99.3 (+1.0) and TTC achieves 99.1 (+1.7). Notably, \mtd{} attains a perfect TL score of 100, demonstrating that the cognitive reward effectively enforces traffic light compliance, a capability that purely physical metrics cannot capture. We note that EP (83.6) and EC (80.0) are lower than some baselines, reflecting a principled safety-progress trade-off consistent with the v1 findings.

\begin{figure*}[!t]
\centering
\includegraphics[width=0.96\textwidth]{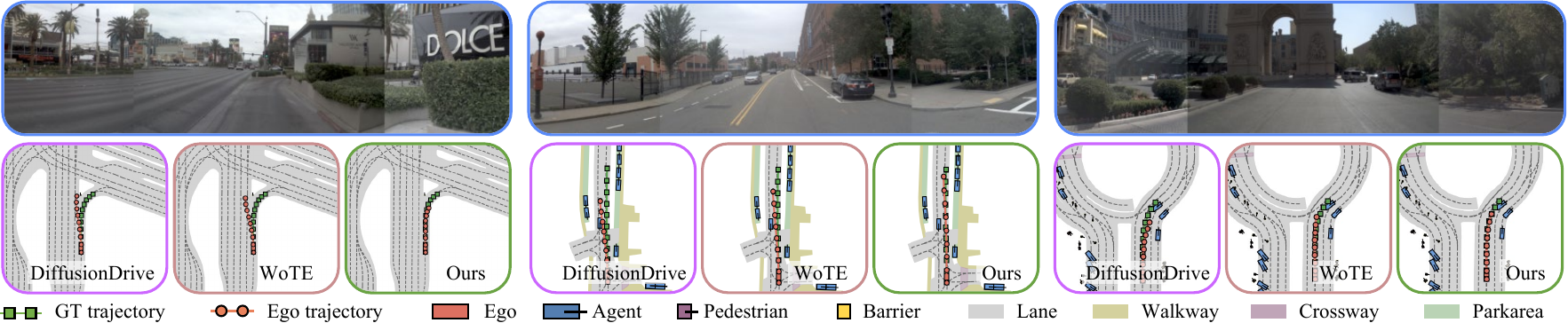}
\vspace{-0.6em}
\caption{Qualitative comparisons with DiffusionDrive~\cite{liao2025ddrive} and WoTE~\cite{li2025end}.
}
\label{fig: compare}
\vspace{-1.3em}
\end{figure*}


\noindent\textbf{Qualitative Comparison.} \cref{fig: compare} validates the holistic advantages of \mtd{} across three representative scenarios. In the leftmost panel, competing methods deviate at a fork, whereas \mtd{} accurately maintains the precise route via the intention-aware \mone{}. The middle panel demonstrates our physical reward derived from world-model rollouts strictly enforcing spatial compliance, effectively preventing the illegal double-yellow line crossings observed in baseline models. Finally, the rightmost panel highlights our physical reward optimizing spatial precision, ensuring rigorous lane centering during complex turns where alternative approaches consistently drift.

\subsection{Ablation Studies}
\label{sec: exp_abl}
We ablate key components on NAVSIM v1 NAVTEST, examining VLM distillation, world model rollout horizon, and the dual cognitive-physical scores/rewards. 
Since the physical score relies on world model rollouts and serves as the foundation for trajectory selection, we adopt this minimal configuration (A0) as the baseline, achieving 87.1 PDMS in \cref{tab:abl1}.

\begin{figure*}[!t]
\centering
\begin{minipage}[t]{0.35\textwidth}
\centering
\includegraphics[width=\textwidth]{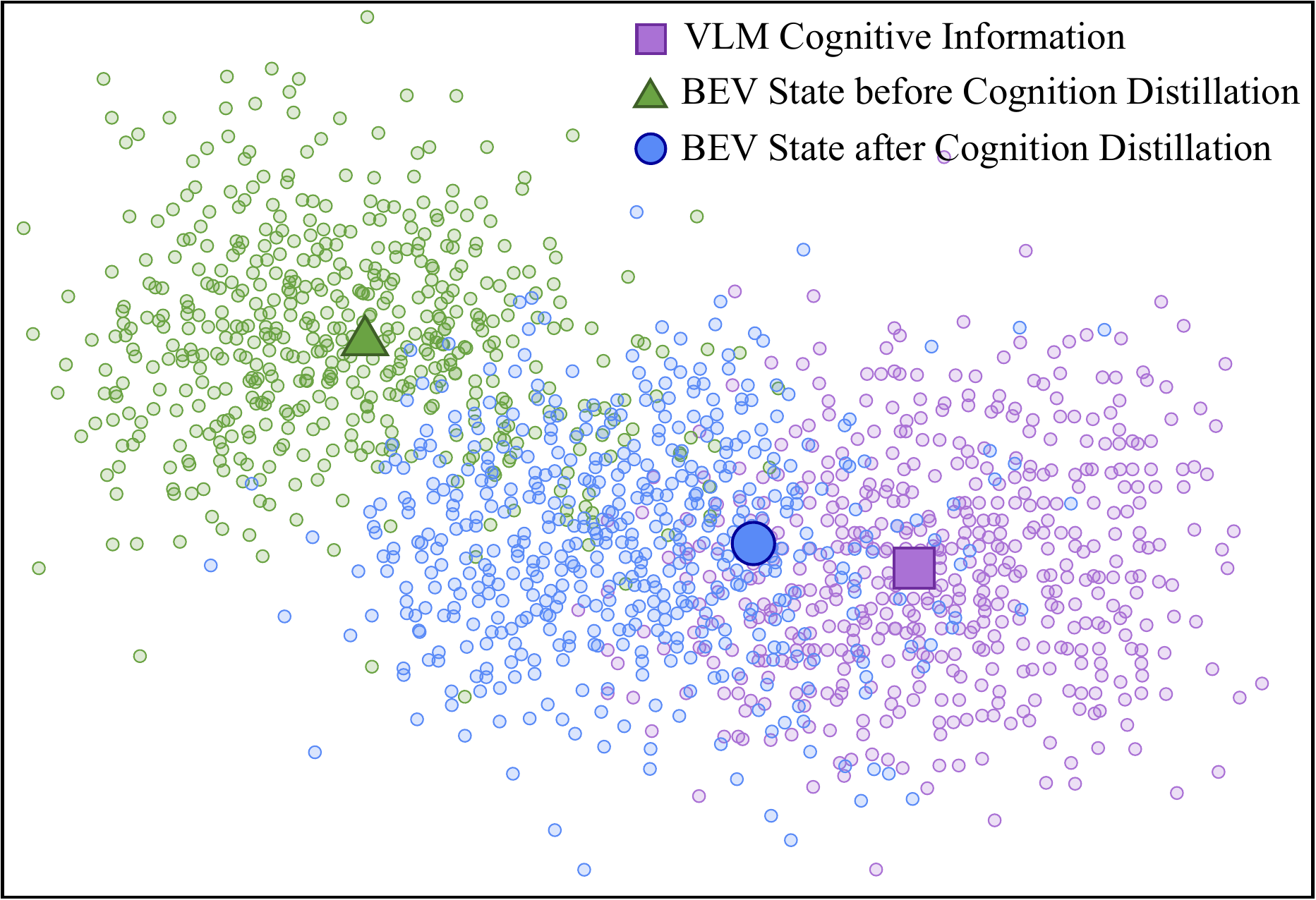}
\vspace{-1.5em}
\caption{t-SNE~\cite{van2008visualizing} of distillation.}
\label{fig: tsne}
\end{minipage}
\hfill
\begin{minipage}[t]{0.6\textwidth}
\centering
\includegraphics[width=\textwidth]{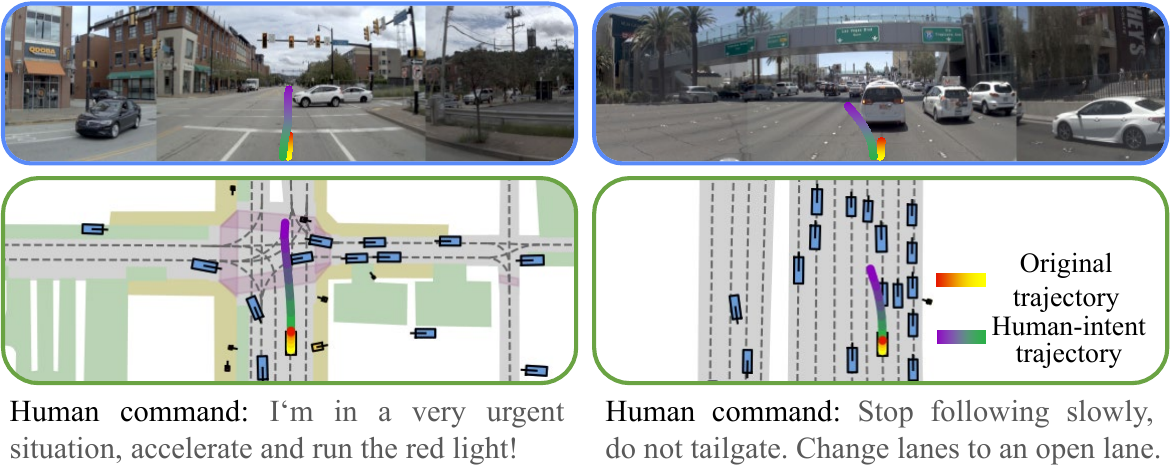}
\vspace{-1.5em}
\caption{Intent-controllable driving via user commands.}
\label{fig: text_control}
\end{minipage}
\vspace{-1.2em}
\end{figure*}

\noindent\textbf{VLM Distillation.}
The goal of VLM distillation (VD) is to transfer cognitive priors from the VLM into the BEV encoder of the driving baseline.
As visualized in \cref{fig: tsne}, before distillation the BEV features exhibit clear separation from VLM cognitive features, whereas after distillation they substantially overlap, confirming effective internalization of cognitive priors.
This knowledge transfer brings notable performance gains: as shown in \cref{tab:abl1}, adding VD (A1) improves PDMS from 87.1 to 88.6 (+1.5), with a notable EP gain of +2.5 and DAC of +1.9.
Moreover, we observe a significant gap of 2.4 PDMS between generic and driving-adapted InternVL3-8B~\cite{zhu2025internvl3} in \cref{tab:ablv}, indicating that domain-specific VLM is essential for autonomous driving.

\noindent\textbf{Hierarchical Trajectory Selection.}
Building on A1 in \cref{tab:abl1}, we add the cognitive score (CS) to form hierarchical trajectory selection that combines both physical and cognitive criteria (A2). This yields PDMS of 88.9 (+0.3), with the most notable gain in TTC (+1.5). 
This is because the cognitive score captures semantic cues such as yielding intentions and traffic priority, enabling earlier anticipation of potential conflicts and thus maintaining safer time margins to other agents.

\noindent\textbf{Cognitive-Physical Policy Optimization.}
Starting from the IL-only configuration (A2), incorporating RL rewards brings the largest performance leap.
CRL alone (B1) contributes +1.3 PDMS, primarily improving NC (+0.5) and EP (+1.0) by aligning with cognitive intentions.
PRL alone (B2) contributes +1.8 PDMS with gains in DAC (+0.6), EP (+1.8), and TTC (+1.3), demonstrating that physical reward effectively guides safer and more efficient driving.
When both rewards are combined (B3), the full model achieves 91.4 PDMS (+2.5 over A2), confirming that the \mthree{} mechanism jointly optimizes physical safety and intent compliance.

\noindent\textbf{Rollout Horizon in World Model.}
The world model rolls out future BEV states over a fixed 4s horizon, with $K$ controlling the number of intermediate steps (i.e., each step spans $4/K$ seconds).
As shown in \cref{tab:ablk}, $K=1$ yields suboptimal safety metrics (DAC 97.2, TTC 95.3), as a single coarse prediction cannot capture mid-trajectory violations.
$K=2$ substantially improves all metrics, achieving the best PDMS (91.4).
Further refining to $K=4$ yields marginal EP gain but degrades NC and TTC, likely due to compounding auto-regressive prediction errors across more intermediate steps.
Thus, $K=2$ strikes the best balance between temporal granularity and error accumulation.

\begin{table*}[!t]
\centering
\begin{minipage}[t]{0.61\textwidth}
\centering
\caption{\small Ablation of key components on NAVSIM v1. VD: VLM distillation, PS/CS: physical/cognitive score, PRL/CRL: RL with physical/cognitive reward.}
\label{tab:abl1}
\resizebox{\textwidth}{!}{
\setlength{\tabcolsep}{3pt}
\begin{tabular}{c|c|cccc|ccccc|c}
\toprule
& PS & VD & CS & CRL & PRL & NC $\uparrow$ & DAC $\uparrow$ & EP $\uparrow$ & TTC $\uparrow$ & C $\uparrow$ & PDMS $\uparrow$ \\
\midrule
A0 & \textcolor{VividGreen}{\ding{51}} &  &  &  &  & 98.2 & 95.7 & 81.1 & 94.2 & \underline{99.9} & 87.1 \\
A1 & \textcolor{VividGreen}{\ding{51}} & \textcolor{VividGreen}{\ding{51}} &  &  &  & 98.1 & 97.6 & 83.6 & 93.0 & \textbf{100} & 88.6 \\
A2 & \textcolor{VividGreen}{\ding{51}} & \textcolor{VividGreen}{\ding{51}} & \textcolor{VividGreen}{\ding{51}} &  &  & 98.3 & 97.2 & 83.2 & 94.5 & \textbf{100} & 88.9 \\
B1 & \textcolor{VividGreen}{\ding{51}} & \textcolor{VividGreen}{\ding{51}} & \textcolor{VividGreen}{\ding{51}} & \textcolor{VividGreen}{\ding{51}} &  & 98.8 & 97.1 & 84.2 & 95.1 & \textbf{100} & 90.2 \\
B2 & \textcolor{VividGreen}{\ding{51}} & \textcolor{VividGreen}{\ding{51}} & \textcolor{VividGreen}{\ding{51}} &  & \textcolor{VividGreen}{\ding{51}} & \underline{98.9} & \underline{97.8} & \underline{85.0} & \underline{95.8} & \textbf{100} & \underline{90.7} \\
\cellcolor{green!7}B3 & \cellcolor{green!7}\textcolor{VividGreen}{\ding{51}} & \cellcolor{green!7}\textcolor{VividGreen}{\ding{51}} & \cellcolor{green!7}\textcolor{VividGreen}{\ding{51}} & \cellcolor{green!7}\textcolor{VividGreen}{\ding{51}} & \cellcolor{green!7}\textcolor{VividGreen}{\ding{51}} & \cellcolor{green!7}\textbf{99.0} & \cellcolor{green!7}\textbf{98.2} & \cellcolor{green!7}\textbf{85.3} & \cellcolor{green!7}\textbf{96.8} & \cellcolor{green!7}\textbf{100} & \cellcolor{green!7}\textbf{91.4} \\
\bottomrule
\end{tabular}%
}
\end{minipage}%
\hfill
\begin{minipage}[t]{0.37\textwidth}
\centering
\caption{\small Effect of VLM fine-tuning.}
\label{tab:ablv}
\resizebox{\textwidth}{!}{
\setlength{\tabcolsep}{4pt}
\begin{tabular}{l|ccccc|c}
\toprule
VLM & NC $\uparrow$ & DAC $\uparrow$ & EP $\uparrow$ & TTC $\uparrow$ & C $\uparrow$ & PDMS $\uparrow$ \\
\midrule
w/o FT & \underline{97.6} & \underline{95.9} & \underline{83.7} & \underline{94.4} & \underline{99.9} & \underline{89.0} \\
\cellcolor{green!7} w/ FT & \cellcolor{green!7}\textbf{99.0} & \cellcolor{green!7}\textbf{98.2} & \cellcolor{green!7}\textbf{85.3} & \cellcolor{green!7}\textbf{96.8} & \cellcolor{green!7}\textbf{100} & \cellcolor{green!7}\textbf{91.4} \\
\bottomrule
\end{tabular}%
}
\vspace{-0.53em}
\caption{\small Effect of rollout horizon $K$ in the world model.}
\label{tab:ablk}
\resizebox{\textwidth}{!}{
\setlength{\tabcolsep}{4pt}
\begin{tabular}{c|ccccc|c}
\toprule
$K$ & NC $\uparrow$ & DAC $\uparrow$ & EP $\uparrow$ & TTC $\uparrow$ & C $\uparrow$ & PDMS $\uparrow$ \\
\midrule
 1 & \underline{98.9} & 97.2 & 84.8 & \underline{95.3} & \textbf{100} & 90.8 \\
\cellcolor{green!7}2 & \cellcolor{green!7}\textbf{99.0} & \cellcolor{green!7}\textbf{98.2} & \cellcolor{green!7}\underline{85.3} & \cellcolor{green!7}\textbf{96.8} & \cellcolor{green!7}\textbf{100} & \cellcolor{green!7}\textbf{91.4} \\
 4 & 98.8 & \underline{98.0} & \textbf{85.4} & 95.1 & \textbf{100} & \underline{91.2} \\

\bottomrule
\end{tabular}%
}
\end{minipage}
\vspace{-0.8em}
\end{table*}

\vspace{-0.2em}
\subsection{Intent-Controllable Driving}
\vspace{-0.2em}
\label{sec: exp_world}
Because VLM distillation releases the cognitive channel at inference time, \mtd{} naturally supports a pluggable intent interface: the distilled feature can be replaced by an encoded user command, steering the policy without retraining.
\cref{fig: text_control} illustrates this capability through two scenarios.
In the left example, the ego vehicle is decelerating for a red light under the default policy. Upon receiving ``\textit{I'm in an urgent situation, accelerate and run the red light!}'', \mtd{} overrides traffic rule compliance and accelerates through the intersection.
In the right example, the ego vehicle is following a slow predecessor. Given ``\textit{Do not tailgate. Change lanes to an open lane.}'', it promptly executes a left lane change.
Notably, during intent-driven execution the world model rollout continues to enforce physical safety constraints, ensuring user commands are fulfilled without collisions. This confirms \mtd{} functions as a controllable yet safe system capable of interpreting diverse human instructions.

\vspace{-0.5em}
\section{Conclusion}
\vspace{-0.5em}

We have presented \textbf{\mtd{}}, a cognitive-physical learning framework for end-to-end autonomous driving. 
By distilling VLM priors into BEV representations, rolling out future states via an explicit world model, and optimizing the policy through GRPO with dual cognitive-physical rewards, \mtd{} has unified semantic understanding, physical safety reasoning, and closed-loop exploration in a single architecture. 
Experiments on NAVSIM v1 and v2 have validated the effectiveness of each component, with \mtd{} achieving state-of-the-art performance across both benchmarks while enabling intent-controllable driving via language commands. 
We hope this work inspires further research on bridging high-level cognition with low-level physical grounding in safety-critical autonomous systems.

\begin{ack}
This work was supported by the National Science Fund of China under Grant Nos. U24A20330, 62361166670, and 62276144.
\end{ack}

\bibliographystyle{plain}
\bibliography{egbib}

\begin{thebibliography}{10}

\bibitem{bai2025qwen3}
Shuai Bai, Yuxuan Cai, Ruizhe Chen, Keqin Chen, Xionghui Chen, Zesen Cheng,
  Lianghao Deng, Wei Ding, Chang Gao, Chunjiang Ge, et~al.
\newblock Qwen3-vl technical report.
\newblock {\em arXiv preprint arXiv:2511.21631}, 2025.

\bibitem{caesar2021nuplan}
Holger Caesar, Juraj Kabzan, Kok~Seang Tan, Whye~Kit Fong, Eric Wolff, Alex
  Lang, Luke Fletcher, Oscar Beijbom, and Sammy Omari.
\newblock nuplan: A closed-loop ml-based planning benchmark for autonomous
  vehicles.
\newblock {\em arXiv preprint arXiv:2106.11810}, 2021.

\bibitem{caopseudo}
Wei Cao, Marcel Hallgarten, Tianyu Li, Daniel Dauner, Xunjiang Gu, Caojun Wang,
  Yakov Miron, Marco Aiello, Hongyang Li, Igor Gilitschenski, Boris Ivanovic,
  Marco Pavone, Andreas Geiger, and Kashyap Chitta.
\newblock Pseudo-simulation for autonomous driving.
\newblock In {\em CoRL}, volume 305, pages 4709--4722. PMLR, 2025.

\bibitem{chen2024end}
Li~Chen, Penghao Wu, Kashyap Chitta, Bernhard Jaeger, Andreas Geiger, and
  Hongyang Li.
\newblock End-to-end autonomous driving: Challenges and frontiers.
\newblock {\em TPAMI}, 46(12):10164--10183, 2024.

\bibitem{chen2024vadv2}
Shaoyu Chen, Bo~Jiang, Hao Gao, Bencheng Liao, Qing Xu, Qian Zhang, Chang
  Huang, Wenyu Liu, and Xinggang Wang.
\newblock Vadv2: End-to-end vectorized autonomous driving via probabilistic
  planning.
\newblock 2026.

\bibitem{chitta2022transfuser}
Kashyap Chitta, Aditya Prakash, Bernhard Jaeger, Zehao Yu, Katrin Renz, and
  Andreas Geiger.
\newblock Transfuser: Imitation with transformer-based sensor fusion for
  autonomous driving.
\newblock {\em TPAMI}, 45(11):12878--12895, 2022.

\bibitem{openscene2023}
OpenScene Contributors.
\newblock Openscene: The largest up-to-date 3d occupancy prediction benchmark
  in autonomous driving.
\newblock \url{https://github.com/OpenDriveLab/OpenScene}, 2023.

\bibitem{dauner2024navsim}
Daniel Dauner, Marcel Hallgarten, Tianyu Li, Xinshuo Weng, Zhiyu Huang, Zetong
  Yang, Hongyang Li, Igor Gilitschenski, Boris Ivanovic, Marco Pavone, et~al.
\newblock Navsim: Data-driven non-reactive autonomous vehicle simulation and
  benchmarking.
\newblock {\em NeurIPS}, pages 28706--28719, 2024.

\bibitem{feng2025artemis}
Renju Feng, Ning Xi, Duanfeng Chu, Rukang Wang, Zejian Deng, Anzheng Wang,
  Liping Lu, Jinxiang Wang, and Yanjun Huang.
\newblock Artemis: Autoregressive end-to-end trajectory planning with mixture
  of experts for autonomous driving.
\newblock {\em RA-L}, 11(1):226--233, 2025.

\bibitem{guo2025deepseek}
Daya Guo, Dejian Yang, Haowei Zhang, Junxiao Song, Peiyi Wang, Qihao Zhu,
  Runxin Xu, Ruoyu Zhang, Shirong Ma, Xiao Bi, et~al.
\newblock {DeepSeek-R1} incentivizes reasoning in {LLMs} through reinforcement
  learning.
\newblock {\em Nature}, 645(8081):633--638, 2025.

\bibitem{hu2022lora}
Edward~J Hu, Yelong Shen, Phillip Wallis, Zeyuan Allen-Zhu, Yuanzhi Li, Shean
  Wang, Liang Wang, Weizhu Chen, et~al.
\newblock Lora: Low-rank adaptation of large language models.
\newblock 2022.

\bibitem{hu2023planning}
Yihan Hu, Jiazhi Yang, Li~Chen, Keyu Li, Chonghao Sima, Xizhou Zhu, Siqi Chai,
  Senyao Du, Tianwei Lin, Wenhai Wang, et~al.
\newblock Planning-oriented autonomous driving.
\newblock In {\em CVPR}, pages 17853--17862, 2023.

\bibitem{hwang2024emma}
Jyh-Jing Hwang, Runsheng Xu, Hubert Lin, Wei-Chih Hung, Jingwei Ji, Kristy
  Choi, Di~Huang, Tong He, Paul Covington, Benjamin Sapp, Yin Zhou, James Guo,
  Dragomir Anguelov, and Mingxing Tan.
\newblock {EMMA}: End-to-end multimodal model for autonomous driving.
\newblock {\em TMLR}, 2025.

\bibitem{jiang2025irl}
Anqing Jiang, Gao Yu, Heng Yuwen, Yiru Wang, Wang Shuo, Jiang Hao, and Sun Hao.
\newblock Irl-vla: Vision-language-action training via reward world model.
\newblock In {\em CVPR Findings}, pages 970--979, 2026.

\bibitem{jiang2024senna}
Bo~Jiang, Shaoyu Chen, Bencheng Liao, Xingyu Zhang, Wei Yin, Qian Zhang, Chang
  Huang, Wenyu Liu, and Xinggang Wang.
\newblock Senna: Bridging large vision-language models and end-to-end
  autonomous driving.
\newblock {\em IJCV}, 134(9):407, 2026.

\bibitem{jiang2023vad}
Bo~Jiang, Shaoyu Chen, Qing Xu, Bencheng Liao, Jiajie Chen, Helong Zhou, Qian
  Zhang, Wenyu Liu, Chang Huang, and Xinggang Wang.
\newblock Vad: Vectorized scene representation for efficient autonomous
  driving.
\newblock In {\em ICCV}, pages 8340--8350, 2023.

\bibitem{jiang2025alphadrive}
Bo~Jiang, Shaoyu Chen, Qian Zhang, Wenyu Liu, and Xinggang Wang.
\newblock Alphadrive: Unleashing the power of vlms in autonomous driving via
  reinforcement learning and reasoning.
\newblock {\em arXiv preprint arXiv:2503.07608}, 2025.

\bibitem{jin2023adapt}
Bu~Jin, Xinyu Liu, Yupeng Zheng, Pengfei Li, Hao Zhao, Tong Zhang, Yuhang
  Zheng, Guyue Zhou, and Jingjing Liu.
\newblock Adapt: Action-aware driving caption transformer.
\newblock In {\em ICRA}, pages 7554--7561, 2023.

\bibitem{li2025finetuning}
Derun Li, Jianwei Ren, Yue Wang, Xin Wen, Pengxiang Li, Leimeng Xu, Kun Zhan,
  Zhongpu Xia, Peng Jia, Xianpeng Lang, et~al.
\newblock Finetuning generative trajectory model with reinforcement learning
  from human feedback.
\newblock {\em arXiv preprint arXiv:2503.10434}, 2025.

\bibitem{li2025hydra++}
Kailin Li, Zhenxin Li, Shiyi Lan, Yuan Xie, Zhizhong Zhang, Jiayi Liu, Zuxuan
  Wu, Zhiding Yu, and Jose~M Alvarez.
\newblock Hydra-mdp++: Advancing end-to-end driving via expert-guided
  hydra-distillation.
\newblock {\em arXiv preprint arXiv:2503.12820}, 2025.

\bibitem{li2025spacedrive}
Peizheng Li, Zhenghao Zhang, David Holtz, Hang Yu, Yutong Yang, Yuzhi Lai, Rui
  Song, Andreas Geiger, and Andreas Zell.
\newblock Spacedrive: Infusing spatial awareness into vlm-based autonomous
  driving.
\newblock In {\em CVPR}, pages 40096--40107, 2026.

\bibitem{li2025enhancing}
Yingyan Li, Lue Fan, Jiawei He, Yuqi Wang, Yuntao Chen, Zhaoxiang Zhang, and
  Tieniu Tan.
\newblock Enhancing end-to-end autonomous driving with latent world model.
\newblock 2025.

\bibitem{li2025drivevla}
Yingyan Li, Shuyao Shang, Weisong Liu, Bing Zhan, Haochen Wang, Yuqi Wang,
  Yuntao Chen, Xiaoman Wang, Yasong An, Chufeng Tang, et~al.
\newblock Drivevla-w0: World models amplify data scaling law in autonomous
  driving.
\newblock {\em ICLR}, 2026.

\bibitem{li2025end}
Yingyan Li, Yuqi Wang, Yang Liu, Jiawei He, Lue Fan, and Zhaoxiang Zhang.
\newblock End-to-end driving with online trajectory evaluation via bev world
  model.
\newblock In {\em ICCV}, pages 27137--27146, 2025.

\bibitem{li2025recogdrive}
Yongkang Li, Kaixin Xiong, Xiangyu Guo, Fang Li, Sixu Yan, Gangwei Xu, Lijun
  Zhou, Long Chen, Haiyang Sun, Bing Wang, et~al.
\newblock Recogdrive: A reinforced cognitive framework for end-to-end
  autonomous driving.
\newblock 2026.

\bibitem{li2026unidrivevla}
Yongkang Li, Lijun Zhou, Sixu Yan, Bencheng Liao, Tianyi Yan, Kaixin Xiong,
  Long Chen, Hongwei Xie, Bing Wang, Guang Chen, et~al.
\newblock Unidrivevla: Unifying understanding, perception, and action planning
  for autonomous driving.
\newblock {\em arXiv preprint arXiv:2604.02190}, 2026.

\bibitem{li2026drive}
Yue Li, Meng Tian, Dechang Zhu, Jiangtong Zhu, Zhenyu Lin, Zhiwei Xiong, and
  Xinhai Zhao.
\newblock Drive-r1: Bridging reasoning and planning in vlms for autonomous
  driving with reinforcement learning.
\newblock In {\em AAAI}, volume~40, pages 6708--6716, 2026.

\bibitem{li2024hydra}
Zhenxin Li, Kailin Li, Shihao Wang, Shiyi Lan, Zhiding Yu, Yishen Ji, Zhiqi Li,
  Ziyue Zhu, Jan Kautz, Zuxuan Wu, et~al.
\newblock Hydra-mdp: End-to-end multimodal planning with multi-target
  hydra-distillation.
\newblock {\em arXiv preprint arXiv:2406.06978}, 2024.

\bibitem{liao2025ddrive}
Bencheng Liao, Shaoyu Chen, Haoran Yin, Bo~Jiang, Cheng Wang, Sixu Yan, Xinbang
  Zhang, Xiangyu Li, Ying Zhang, Qian Zhang, et~al.
\newblock Diffusiondrive: Truncated diffusion model for end-to-end autonomous
  driving.
\newblock In {\em CVPR}, pages 12037--12047, 2025.

\bibitem{lin2017focal}
Tsung-Yi Lin, Priya Goyal, Ross Girshick, Kaiming He, and Piotr Doll{\'a}r.
\newblock Focal loss for dense object detection.
\newblock In {\em ICCV}, pages 2980--2988, 2017.

\bibitem{liu2025x}
Wei Liu, Jiyuan Zhang, Binxiong Zheng, Yufeng Hu, Yingzhan Lin, and Zengfeng
  Zeng.
\newblock X-driver: Explainable autonomous driving with vision-language models.
\newblock {\em arXiv preprint arXiv:2505.05098}, 2025.

\bibitem{loshchilov2017sgdr}
Ilya Loshchilov and Frank Hutter.
\newblock Sgdr: Stochastic gradient descent with warm restarts.
\newblock In {\em ICLR}, 2017.

\bibitem{loshchilov2017decoupled}
Ilya Loshchilov and Frank Hutter.
\newblock Decoupled weight decay regularization.
\newblock 2018.

\bibitem{luo2025adathinkdrive}
Yuechen Luo, Fang Li, Shaoqing Xu, Zhiyi Lai, Lei Yang, Qimao Chen, Ziang Luo,
  Zixun Xie, Shengyin Jiang, Jiaxin Liu, et~al.
\newblock Adathinkdrive: Adaptive thinking via reinforcement learning for
  autonomous driving.
\newblock In {\em ICRA}, pages 14489--14496, 2026.

\bibitem{malla2023drama}
Srikanth Malla, Chiho Choi, Isht Dwivedi, Joon~Hee Choi, and Jiachen Li.
\newblock Drama: Joint risk localization and captioning in driving.
\newblock In {\em WACV}, pages 1043--1052, 2023.

\bibitem{marcu2024lingoqa}
Ana-Maria Marcu, Long Chen, Jan H{\"u}nermann, Alice Karnsund, Benoit Hanotte,
  Prajwal Chidananda, Saurabh Nair, Vijay Badrinarayanan, Alex Kendall, Jamie
  Shotton, et~al.
\newblock Lingoqa: Visual question answering for autonomous driving.
\newblock In {\em ECCV}, pages 252--269. Springer, 2024.

\bibitem{oord2018representation}
Aaron van~den Oord, Yazhe Li, and Oriol Vinyals.
\newblock Representation learning with contrastive predictive coding.
\newblock {\em arXiv preprint arXiv:1807.03748}, 2018.

\bibitem{park2024vlaad}
SungYeon Park, MinJae Lee, JiHyuk Kang, Hahyeon Choi, Yoonah Park, Juhwan Cho,
  Adam Lee, and DongKyu Kim.
\newblock Vlaad: Vision and language assistant for autonomous driving.
\newblock In {\em WACV Workshops}, pages 980--987, 2024.

\bibitem{prakash2021multi}
Aditya Prakash, Kashyap Chitta, and Andreas Geiger.
\newblock Multi-modal fusion transformer for end-to-end autonomous driving.
\newblock In {\em CVPR}, pages 7077--7087, 2021.

\bibitem{qian2024nuscenes}
Tianwen Qian, Jingjing Chen, Linhai Zhuo, Yang Jiao, and Yu-Gang Jiang.
\newblock Nuscenes-qa: A multi-modal visual question answering benchmark for
  autonomous driving scenario.
\newblock In {\em AAAI}, volume~38, pages 4542--4550, 2024.

\bibitem{radford2021learning}
Alec Radford, Jong~Wook Kim, Chris Hallacy, Aditya Ramesh, Gabriel Goh,
  Sandhini Agarwal, Girish Sastry, Amanda Askell, Pamela Mishkin, Jack Clark,
  et~al.
\newblock Learning transferable visual models from natural language
  supervision.
\newblock In {\em ICML}, pages 8748--8763. PMLR, 2021.

\bibitem{renz2025simlingo}
Katrin Renz, Long Chen, Elahe Arani, and Oleg Sinavski.
\newblock Simlingo: Vision-only closed-loop autonomous driving with
  language-action alignment.
\newblock In {\em CVPR}, pages 11993--12003, 2025.

\bibitem{rowe2025poutine}
Luke Rowe, Rodrigue de~Schaetzen, Roger Girgis, Christopher Pal, and Liam
  Paull.
\newblock Poutine: Vision-language-trajectory pre-training and reinforcement
  learning post-training enable robust end-to-end autonomous driving.
\newblock {\em arXiv preprint arXiv:2506.11234}, 2025.

\bibitem{shang2025drivedpo}
Shuyao Shang, Yuntao Chen, Yuqi Wang, Yingyan Li, and ZHAO-XIANG ZHANG.
\newblock Drivedpo: Policy learning via safety dpo for end-to-end autonomous
  driving.
\newblock {\em NeurIPS}, pages 81565--81585, 2026.

\bibitem{shao2024lmdrive}
Hao Shao, Yuxuan Hu, Letian Wang, Guanglu Song, Steven~L Waslander, Yu~Liu, and
  Hongsheng Li.
\newblock Lmdrive: Closed-loop end-to-end driving with large language models.
\newblock In {\em CVPR}, pages 15120--15130, 2024.

\bibitem{shao2024deepseekmath}
Zhihong Shao, Peiyi Wang, Qihao Zhu, Runxin Xu, Junxiao Song, Xiao Bi, Haowei
  Zhang, Mingchuan Zhang, YK~Li, Yang Wu, et~al.
\newblock Deepseekmath: Pushing the limits of mathematical reasoning in open
  language models.
\newblock {\em arXiv preprint arXiv:2402.03300}, 2024.

\bibitem{sheng2026explorevla}
Zihao Sheng, Xin Ye, Jingru Luo, Sikai Chen, and Liu Ren.
\newblock Explorevla: Dense world modeling and exploration for end-to-end
  autonomous driving.
\newblock In {\em ECCV}, pages 266--284. Springer, 2026.

\bibitem{sima2024drivelm}
Chonghao Sima, Katrin Renz, Kashyap Chitta, Li~Chen, Hanxue Zhang, Chengen Xie,
  Jens Bei{\ss}wenger, Ping Luo, Andreas Geiger, and Hongyang Li.
\newblock Drivelm: Driving with graph visual question answering.
\newblock In {\em ECCV}, pages 256--274. Springer, 2024.

\bibitem{tang2025deep}
Chen Tang, Ben Abbatematteo, Jiaheng Hu, Rohan Chandra, Roberto
  Mart{\'\i}n-Mart{\'\i}n, and Peter Stone.
\newblock Deep reinforcement learning for robotics: A survey of real-world
  successes.
\newblock {\em Annual Review of Control, Robotics, and Autonomous Systems},
  8(1):153--188, 2025.

\bibitem{van2008visualizing}
Laurens Van~der Maaten and Geoffrey Hinton.
\newblock Visualizing data using t-sne.
\newblock {\em JMLR}, 9(11), 2008.

\bibitem{vaswani2017attention}
Ashish Vaswani, Noam Shazeer, Niki Parmar, Jakob Uszkoreit, Llion Jones,
  Aidan~N Gomez, {\L}ukasz Kaiser, and Illia Polosukhin.
\newblock Attention is all you need.
\newblock {\em NeurIPS}, 2017.

\bibitem{wang2025alpamayo}
Yan Wang, Wenjie Luo, Junjie Bai, Yulong Cao, Tong Che, Ke~Chen, Yuxiao Chen,
  Jenna Diamond, Yifan Ding, Wenhao Ding, et~al.
\newblock Alpamayo-r1: Bridging reasoning and action prediction for
  generalizable autonomous driving in the long tail.
\newblock {\em arXiv preprint arXiv:2511.00088}, 2025.

\bibitem{weng2024drive}
Xinshuo Weng, Boris Ivanovic, Yan Wang, Yue Wang, and Marco Pavone.
\newblock Para-drive: Parallelized architecture for real-time autonomous
  driving.
\newblock In {\em CVPR}, pages 15449--15458, 2024.

\bibitem{wu2026gem}
Yang Wu, Zhaojiang Liu, Qiang Meng, Youquan Liu, Renliang Weng, Jianjun Qian,
  Jian Yang, and Jin Xie.
\newblock Gem: Generating lidar world model via deformable mamba.
\newblock In {\em CVPR}, pages 24227--24236, 2026.

\bibitem{wu2024text2lidar}
Yang Wu, Kaihua Zhang, Jianjun Qian, Jin Xie, and Jian Yang.
\newblock Text2lidar: Text-guided lidar point cloud generation via
  equirectangular transformer.
\newblock In {\em ECCV}, pages 291--310. Springer, 2024.

\bibitem{wu2025weathergen}
Yang Wu, Yun Zhu, Kaihua Zhang, Jianjun Qian, Jin Xie, and Jian Yang.
\newblock Weathergen: A unified diverse weather generator for lidar point
  clouds via spider mamba diffusion.
\newblock In {\em CVPR}, pages 17019--17028, 2025.

\bibitem{xia2026drivelaw}
Tianze Xia, Yongkang Li, Lijun Zhou, Jingfeng Yao, Kaixin Xiong, Haiyang Sun,
  Bing Wang, Kun Ma, Guang Chen, Hangjun Ye, et~al.
\newblock Drivelaw: Unifying planning and video generation in a latent driving
  world.
\newblock {\em CVPR}, 2026.

\bibitem{xing2025goalflow}
Zebin Xing, Xingyu Zhang, Yang Hu, Bo~Jiang, Tong He, Qian Zhang, Xiaoxiao
  Long, and Wei Yin.
\newblock Goalflow: Goal-driven flow matching for multimodal trajectories
  generation in end-to-end autonomous driving.
\newblock In {\em CVPR}, pages 1602--1611, 2025.

\bibitem{xu2025wam}
Mingwang Xu, Jiahao Cui, Feipeng Cai, Hanlin Shang, Zhihao Zhu, Shan Luan,
  Yifang Xu, Neng Zhang, Yaoyi Li, Jia Cai, et~al.
\newblock Wam-diff: A masked diffusion vla framework with moe and online
  reinforcement learning for autonomous driving.
\newblock {\em arXiv preprint arXiv:2512.11872}, 2025.

\bibitem{xu2024drivegpt4}
Zhenhua Xu, Yujia Zhang, Enze Xie, Zhen Zhao, Yong Guo, Kwan-Yee~K Wong,
  Zhenguo Li, and Hengshuang Zhao.
\newblock Drivegpt4: Interpretable end-to-end autonomous driving via large
  language model.
\newblock {\em RA-L}, 9(10):8186--8193, 2024.

\bibitem{yan2025ad}
Tianyi Yan, Tao Tang, Xingtai Gui, Yongkang Li, Jiasen Zheng, Weiyao Huang,
  Lingdong Kong, Wencheng Han, Xia Zhou, Xueyang Zhang, et~al.
\newblock Ad-r1: Closed-loop reinforcement learning for end-to-end autonomous
  driving with impartial world models.
\newblock In {\em CVPR Findings}, pages 1085--1095, 2026.

\bibitem{yang2025raw2drive}
Zhenjie Yang, Xiaosong Jia, Qifeng Li, Xue Yang, Maoqing Yao, and Junchi Yan.
\newblock Raw2drive: Reinforcement learning with aligned world models for
  end-to-end autonomous driving (in carla v2).
\newblock {\em NeurIPS}, pages 134122--134147, 2026.

\bibitem{yao2026drivesuprim}
Wenhao Yao, Zhenxin Li, Shiyi Lan, Zi~Wang, Xinglong Sun, Jose~M Alvarez, and
  Zuxuan Wu.
\newblock Drivesuprim: Towards precise trajectory selection for end-to-end
  planning.
\newblock In {\em AAAI}, volume~40, pages 11910--11918, 2026.

\bibitem{yuan2024drama}
Chengran Yuan, Zhanqi Zhang, Jiawei Sun, Shuo Sun, Zefan Huang, Christina
  Dao~Wen Lee, Dongen Li, Yuhang Han, Anthony Wong, Keng~Peng Tee, et~al.
\newblock Drama: An efficient end-to-end motion planner for autonomous driving
  with mamba.
\newblock {\em arXiv preprint arXiv:2408.03601}, 2024.

\bibitem{zhang2025epona}
Kaiwen Zhang, Zhenyu Tang, Xiaotao Hu, Xingang Pan, Xiaoyang Guo, Yuan Liu,
  Jingwei Huang, Li~Yuan, Qian Zhang, Xiao-Xiao Long, et~al.
\newblock Epona: Autoregressive diffusion world model for autonomous driving.
\newblock In {\em ICCV}, pages 27220--27230, 2025.

\bibitem{zhao2025survey}
Jingyuan Zhao, Yuyan Wu, Rui Deng, Susu Xu, Jinpeng Gao, and Andrew Burke.
\newblock A survey of autonomous driving from a deep learning perspective.
\newblock {\em ACM Computing Surveys}, 57(10):1--60, 2025.

\bibitem{zheng2024genad}
Wenzhao Zheng, Ruiqi Song, Xianda Guo, Chenming Zhang, and Long Chen.
\newblock Genad: Generative end-to-end autonomous driving.
\newblock In {\em ECCV}, pages 87--104. Springer, 2024.

\bibitem{zheng2025world4drive}
Yupeng Zheng, Pengxuan Yang, Zebin Xing, Qichao Zhang, Yuhang Zheng, Yinfeng
  Gao, Pengfei Li, Teng Zhang, Zhongpu Xia, Peng Jia, et~al.
\newblock World4drive: End-to-end autonomous driving via intention-aware
  physical latent world model.
\newblock In {\em ICCV}, pages 28632--28642, 2025.

\bibitem{zheng2025resad}
Zhiyu Zheng, Shaoyu Chen, Haoran Yin, Xinbang Zhang, Jialv Zou, Xinggang Wang,
  Qian Zhang, and Lefei Zhang.
\newblock Resad: Normalized residual trajectory modeling for end-to-end
  autonomous driving.
\newblock {\em CVPR}, 2026.

\bibitem{zhou2026opendrivevla}
Xingcheng Zhou, Xuyuan Han, Feng Yang, Yunpu Ma, Volker Tresp, and Alois Knoll.
\newblock Opendrivevla: Towards end-to-end autonomous driving with large vision
  language action model.
\newblock In {\em AAAI}, volume~40, pages 13782--13790, 2026.

\bibitem{zhou2025autovla}
Zewei Zhou, Tianhui Cai, Seth Zhao, Yun Zhang, Zhiyu Huang, Bolei Zhou, and
  Jiaqi Ma.
\newblock Autovla: A vision-language-action model for end-to-end autonomous
  driving with adaptive reasoning and reinforcement fine-tuning.
\newblock {\em NeurIPS}, pages 27920--27956, 2026.

\bibitem{zhu2025internvl3}
Jinguo Zhu, Weiyun Wang, Zhe Chen, Zhaoyang Liu, Shenglong Ye, Lixin Gu, Hao
  Tian, Yuchen Duan, Weijie Su, Jie Shao, et~al.
\newblock Internvl3: Exploring advanced training and test-time recipes for
  open-source multimodal models.
\newblock {\em arXiv preprint arXiv:2504.10479}, 2025.

\end{thebibliography}

\clearpage
\appendix

\section{Technical appendices and supplementary material}
\label{sec: appen}
This appendix presents additional architectural details, extended experiments, and further analyses.
\cref{sec: app_arch} details the internal network architecture of each core module.
\cref{sec: app_threshold} evaluates the sensitivity of the cognitive threshold $\psi$ within the cognitive and physical trajectory scores.
\cref{sec: app_intent} showcases additional visual demonstrations of human-intent trajectories.
\cref{sec: app_compare} provides extensive qualitative comparisons against DiffusionDrive~\cite{liao2025ddrive} and WoTE~\cite{li2025end}.
\cref{sec: app_traj_num} ablates the number of the candidate trajectories.
\cref{sec: app_efficiency} assesses the computational efficiency and deployment feasibility of our framework.
\cref{sec: app_vlm} provides comparisons with Vision-Language-Action (VLA) methods and the distillation across various VLM backbones.
\cref{sec: app_bev} visualizes the future BEV semantic map predicted by our world model.
\cref{sec: person} demonstrates how customizable driving behaviors can be achieved by modulating the physical reward coefficients.
\cref{sec: app_limit} concludes with a discussion on the inherent limitations and broader impacts of our approach.

\subsection{More Architecture Details}
\label{sec: app_arch}

\begin{figure*}[!h]
\includegraphics[width=1.0\textwidth]{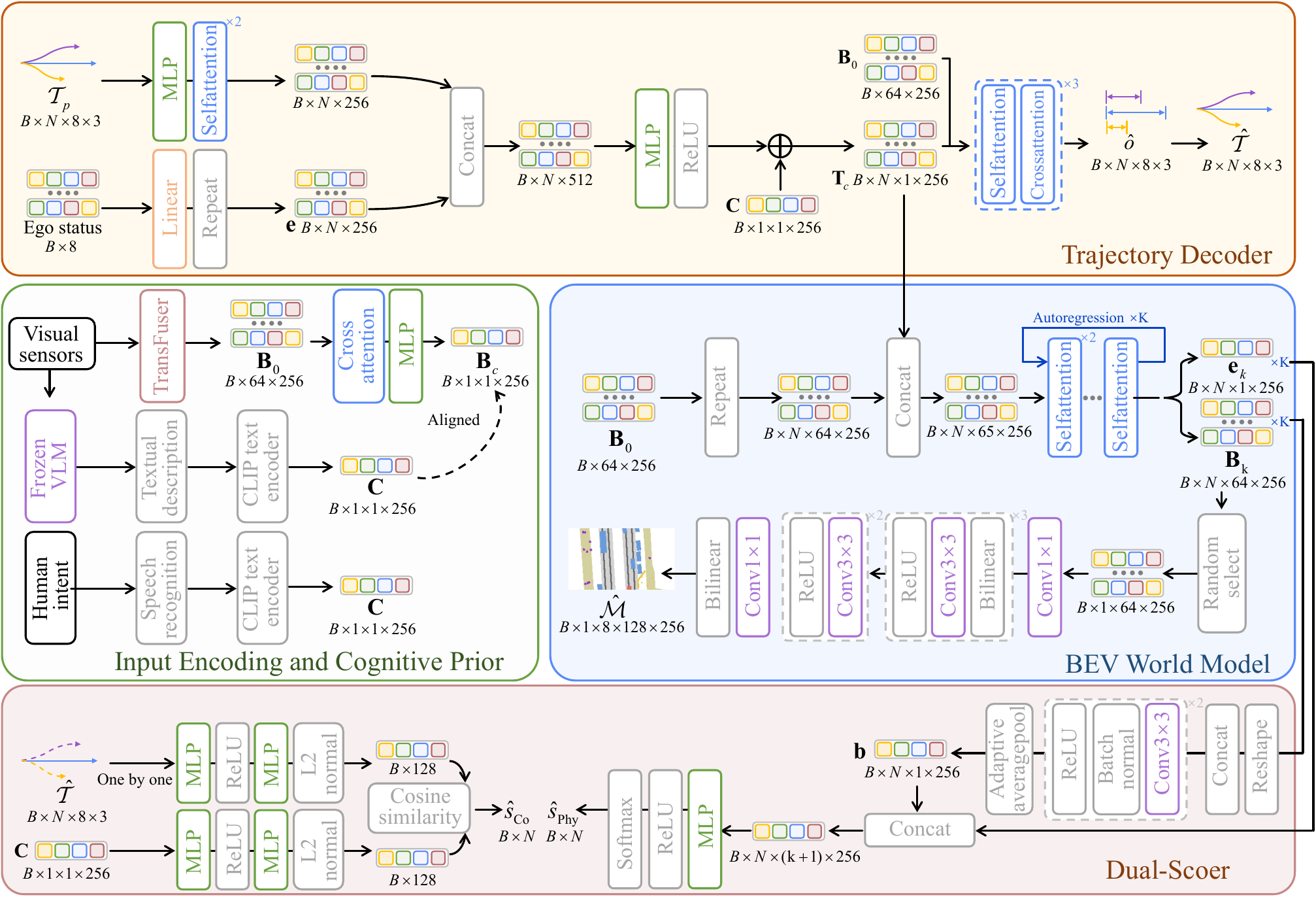}
\vspace{-1.5em}
\caption{Detailed architecture of trajectory decoder and BEV world model in our method.}
\label{fig: app_arch}
\end{figure*}


\paragraph{Input Encoding and Cognitive Feature.}
The model takes a front-view RGB image $\mathcal{V}_{\text{RGB}} \in \mathbb{R}^{B \times 3 \times 256 \times 1024}$ and a surround-view LiDAR BEV $\mathcal{V}_{\text{LiDAR}} \in \mathbb{R}^{B \times 1 \times 256 \times 256}$ as input. Both modalities are jointly processed by a ResNet-34-based TransFuser backbone~\cite{chitta2022transfuser} to yield the foundational BEV state $\textbf{B}_0 \in \mathbb{R}^{B \times 64 \times 256}$.
To construct the cognitive feature, a frozen VLM processes the front-view RGB to generate descriptive text, which is subsequently embedded by a frozen CLIP text encoder~\cite{radford2021learning} to yield $\textbf{C} \in \mathbb{R}^{B \times 1 \times 1 \times 256}$.

Crucially, at inference, the heavy VLM is entirely discarded. When a human-intent command is provided, the text is passed through the CLIP text encoder to produce $\textbf{C}$. In the default autonomous mode (i.e., no explicit command), $\textbf{C}$ is seamlessly replaced by the distilled BEV state $\textbf{B}_{c}$, which is projected via multi-head attention to match the target dimension $\mathbb{R}^{B \times 1 \times 1 \times 256}$. Since $\textbf{B}_{c}$ has already been explicitly aligned with the textual semantic space during pre-training, it serves as a faithful substitute for $\textbf{C}$ without incurring any VLM inference overhead.


\paragraph{Trajectory Decoder.}
The trajectory decoder fuses ego dynamics, kinematic prototypes, and cognitive priors to generate intention-aware candidate trajectories.
First, the ego status, consisting of the driving command $\in \mathbb{R}^{B \times 4}$, ego velocity $\in \mathbb{R}^{B \times 2}$, and ego acceleration $\in \mathbb{R}^{B \times 2}$, is projected via a linear layer and expanded to form the dense ego status feature $\textbf{e} \in \mathbb{R}^{B \times N \times 256}$.
Concurrently, the pre-clustered trajectory prototypes $\mathcal{T}_{p} \in \mathbb{R}^{B \times N \times 8 \times 3}$ are encoded via an MLP followed by a self-attention module to extract the prototype feature.
These ego and prototype features are concatenated, fused through an MLP, and subsequently combined via element-wise addition with the cognitive feature $\textbf{C}$. This explicit injection yields the cognitive trajectory feature $\textbf{T}_{c} \in \mathbb{R}^{B \times N \times 1 \times 256}$.
Finally, $\textbf{T}_{c}$ is decoded through a decoder consisting of self-attention and cross-attention layers: self-attention first models inter-trajectory interactions among the $N$ candidates, then the refined feature serves as queries in cross-attention with $\textbf{B}_0$ as keys and values to aggregate spatial context from the BEV state. The decoded feature is projected via an MLP to predict the residual coordinate offsets $\hat{o} \in \mathbb{R}^{B \times N \times 8 \times 3}$, and the final candidate trajectories are obtained as $\hat{\mathcal{T}} = \mathcal{T}_{p} + \hat{o}$.


\paragraph{BEV World Model.}
The world model auto-regressively predicts future BEV states conditioned on the current spatial observations, candidate actions, and cognitive priors.
The current BEV state $\textbf{B}_{0} \in \mathbb{R}^{B \times 64 \times 256}$ is initially expanded across the trajectory dimension to form a tensor of shape $\mathbb{R}^{B \times N \times 64 \times 256}$. This tensor is subsequently concatenated with the cognitive trajectory feature along the channel dimension, yielding a joint representation $\in \mathbb{R}^{B \times N \times 65 \times 256}$.
This joint feature is processed by two successive self-attention layers to predict the next-step ego feature $\{\textbf{e}_{k}\}_{k=1}^{K}$ and the future BEV state $\{\textbf{B}_{k}\}_{k=1}^{K}$. This predictive process auto-regressively unrolls up to the 4\,s temporal horizon.
To mitigate computational overhead during training, we randomly sample a single trajectory branch, isolating its full 4\,s BEV state evolution $\in \mathbb{R}^{B \times 1 \times 64 \times 256}$.
The sampled state is then projected into the dense future BEV semantic map $\mathcal{\hat{M}} \in \mathbb{R}^{B \times 1 \times 8 \times 128 \times 256}$ through two sequential modules. The first is an upsampling head comprising three cascaded blocks of bilinear upsampling, $3{\times}3$ convolutions, and ReLU activations. The second is a semantic decoding head built from two successive $3{\times}3$ convolution and ReLU blocks, capped by a $1{\times}1$ convolutional layer and a final bilinear upsampling operation.


\paragraph{Hierarchical Trajectory Selection.}
The trajectory selection module evaluates each candidate trajectory along independent physical and cognitive axes.
For the physical score, the $K$-step auto-regressive rollouts of the BEV world model provide the predicted BEV states $\{\textbf{B}_{k}\}_{k=1}^{K}$ and ego features $\{\textbf{e}_{k}\}_{k=1}^{K}$. The BEV states are concatenated along the temporal dimension and processed via convolutional layers and adaptive average pooling, yielding a compact BEV vector $\textbf{b} \in \mathbb{R}^{B \times N \times 1 \times 256}$. This spatial vector is subsequently concatenated with the $K$-step ego features $\{\textbf{e}_{k}\}_{k=1}^{K}$ and passed through an MLP and a softmax activation, producing the final physical scores $\hat{s}_{\texttt{Phy}} \in \mathbb{R}^{B \times N}$ across all candidate trajectories.
For the cognitive score, we independently compute the alignment metric $\hat{s}_{\texttt{Co}}$ for the predicted candidate trajectories $\hat{\mathcal{T}}$. Specifically, an individual candidate $\hat{\mathcal{T}}^{i}$ is projected through an MLP and $L_2$-normalized to extract a trajectory feature vector $\in \mathbb{R}^{B \times 128}$. The cognitive feature $\textbf{C}$ undergoes the exact same MLP mapping and $L_2$ normalization to produce a cognitive vector of identical dimensions. Iterating this alignment process over every candidate yields the dense cognitive scores $\hat{s}_{\texttt{Co}} \in \mathbb{R}^{B \times N}$, which mathematically represent the cosine similarity between the two normalized representations.

\subsection{Cognitive Threshold Sensitivity}
\label{sec: app_threshold}

\begin{figure*}[!h]
\includegraphics[width=1.0\textwidth]{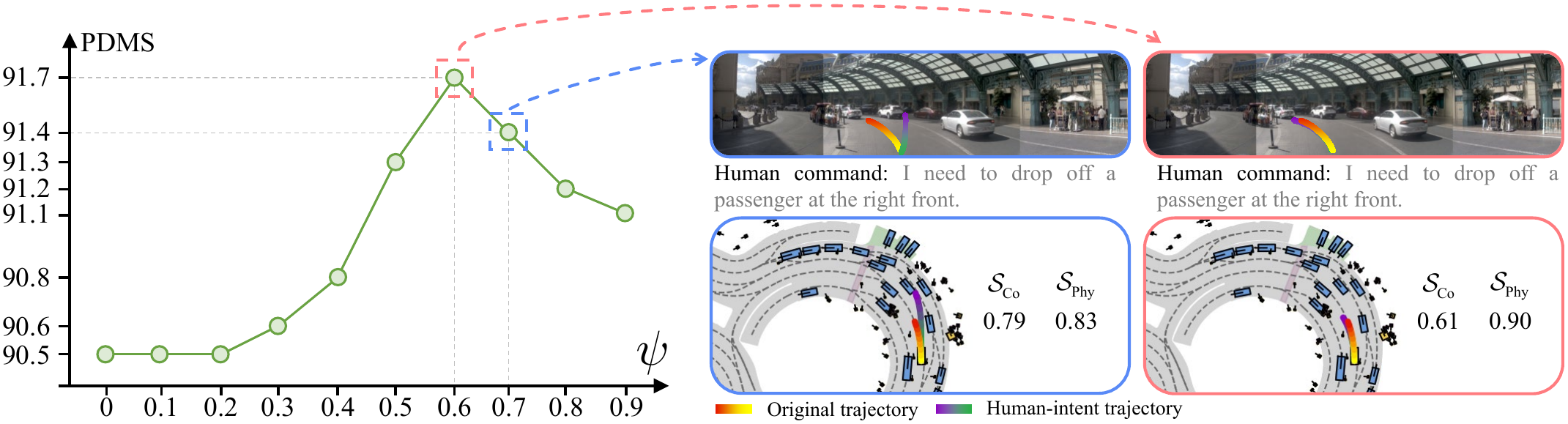}
\vspace{-1.5em}
\caption{Sensitivity analysis of the cognitive threshold $\psi$.}
\label{fig: app_threshold}
\end{figure*}

As illustrated in the left panel of \cref{fig: app_threshold}, increasing $\psi$ progressively enhances the quality of the filtered candidates. A stricter threshold enforces stronger cognitive priors, equipping the model to better decode traffic signals and navigate complex environments. Although pure driving metrics peak at $\psi=0.6$ before declining, we empirically adopt $\psi=0.7$ to secure the optimal equilibrium between physical driving safety and strict compliance with human intents.

Complementing this, the right panel of \cref{fig: app_threshold} visualizes the thresholding dynamics. Under a lenient threshold of $\psi=0.6$, a candidate with a marginal cognitive score of $\hat{s}_{\texttt{Co}}=0.61$ might survive the filter and be selected solely due to its superior physical score $\hat{s}_{\texttt{Phy}}$, despite blatantly violating the human command. By elevating the threshold to $\psi=0.7$, this non-compliant candidate is effectively pruned, strictly guaranteeing the selection of an intention-aligned trajectory.

\subsection{More Human-Intent Driving Visualizations}
\label{sec: app_intent}

\begin{figure*}[!h]
\includegraphics[width=1.0\textwidth]{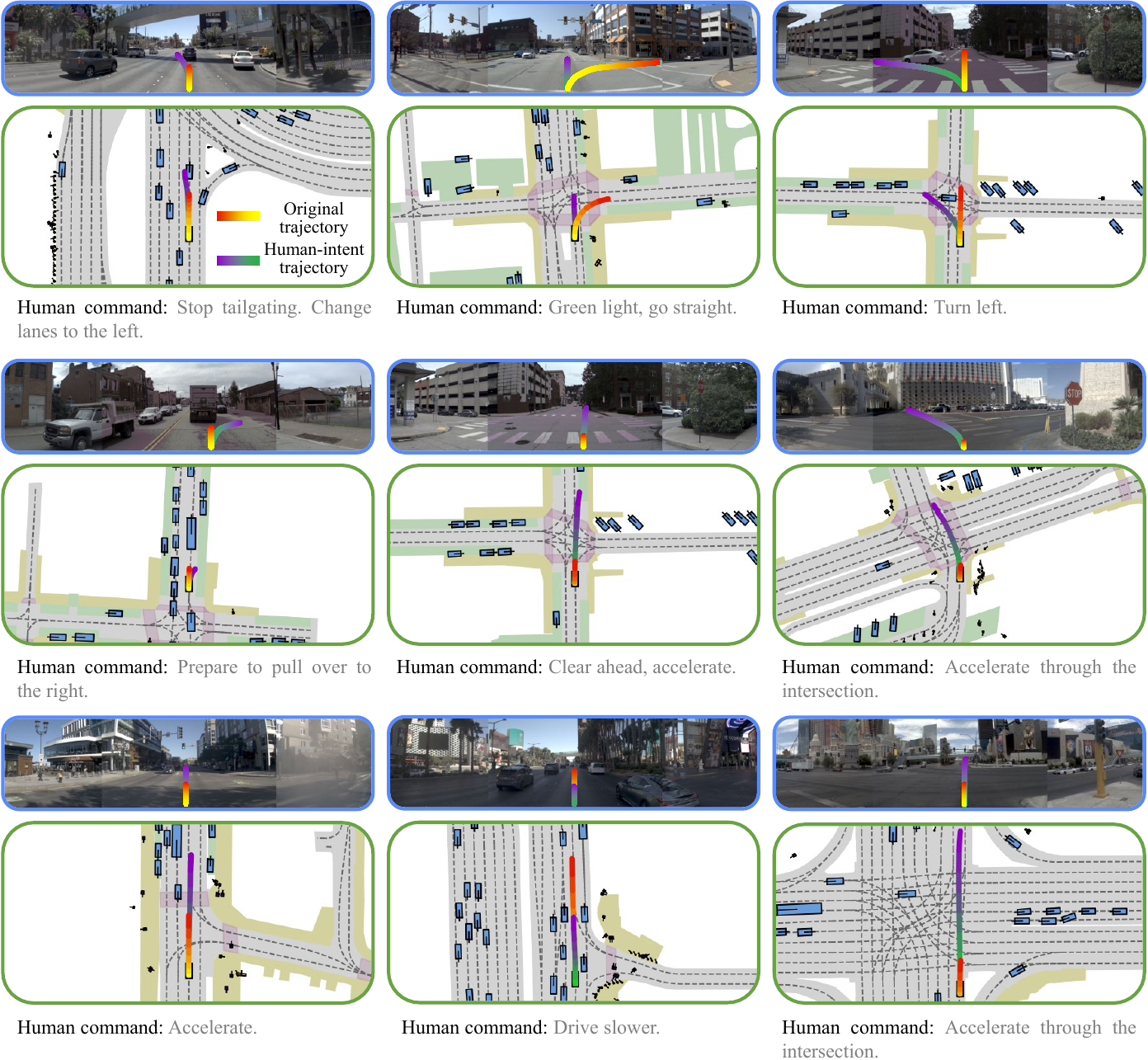}
\vspace{-1.5em}
\caption{More human-intent driving visualizations.}
\label{fig: app_intent}
\end{figure*}

To further validate our cognitive alignment, \cref{fig: app_intent} presents diverse human-guided driving scenarios across three distinct categories.
The first row showcases direction adjustments, encompassing lane changes and turning maneuvers. For instance, in the second example, the ego vehicle dynamically overrides its initial physical plan to proceed straight through a green light, seamlessly executing a right turn solely based on the cognitive command.

The second row highlights start-and-stop maneuvers, such as transitioning from car-following to pulling over. Notably, in the third example, despite the presence of a STOP sign dictating a halt, the vehicle strictly complies with a human prompt to accelerate through the intersection. This deliberate rule-breaking underscores a crucial capability: our policy faithfully understands and prioritizes high-level cognitive intents over rigid physical rules when explicitly instructed.

Finally, the third row depicts speed modulations. In the second example, while the physical environment permits high-speed cruising, a human driver might feel unsafe amidst dense traffic. Upon receiving a deceleration command, the vehicle correctly adopts a cautious profile and slows down, perfectly mirroring human driving etiquette.

\subsection{More Qualitative Comparisons}
\label{sec: app_compare}

\begin{figure*}[!h]
\includegraphics[width=1.0\textwidth]{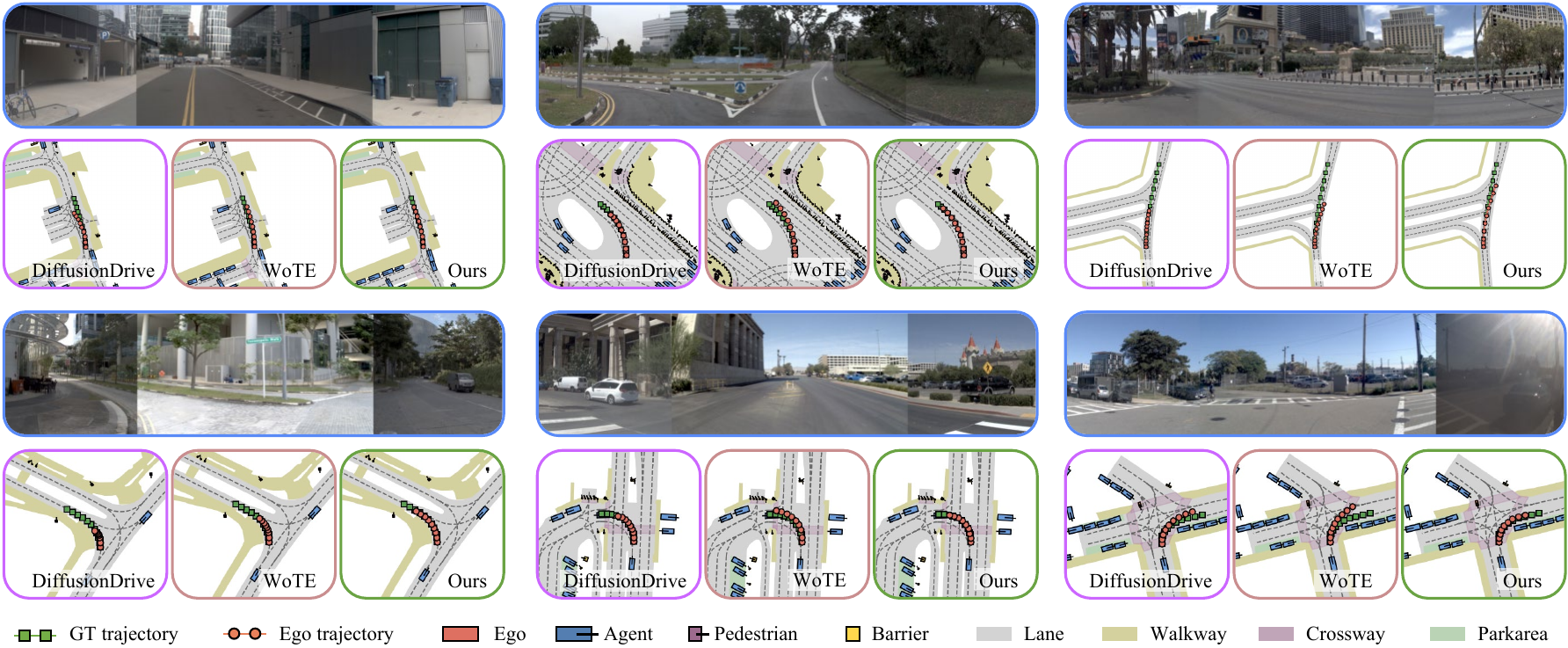}
\vspace{-1.5em}
\caption{More qualitative comparison results.}
\label{fig: app_compare}
\end{figure*}

To further demonstrate our framework's superiority, \cref{fig: app_compare} presents qualitative comparisons against popular leading open-source methods~\cite{liao2025ddrive, li2025end}. 
The first row illustrates straight-line and merging scenarios. In the first example, competing baselines illegally cross the double yellow line or encroach upon hatched markings. In contrast, our method maintains strict lane adherence, enforced by the physical constraints derived from world-model rollouts. 
The second narrow-road example reveals that baselines suffer from sluggish ego progress or trajectory deviations, whereas our policy executes efficient and stable lane-keeping. 
In the third merging scenario, while our method exhibits a minor trajectory adjustment, it safely remains within the drivable area and progresses smoothly. Conversely, the baselines either drift entirely off-road or erroneously halt at the subsequent fork.

The second row showcases complex left-turn and right-turn maneuvers. Although turning inherently limits the overall ego progress across all methods, ours sustains a noticeably higher speed profile. Benefiting from the seamless integration of high-level cognitive awareness and low-level physical constraints, our agent successfully navigates the intersections without any lane departure.

\subsection{Impact of Trajectory Quantity}
\label{sec: app_traj_num}

\begin{table}[!h]
\centering
\caption{Impact of the number of candidate trajectories $N$ on NAVSIM NAVTEST v1.}
\label{tab: app_traj_num}
\resizebox{0.57\textwidth}{!}{
\begin{tabular}{c|ccccc|c|c}
\toprule
$N$ & NC $\uparrow$ & DAC $\uparrow$ & EP $\uparrow$ & TTC $\uparrow$ & C $\uparrow$ & PDMS $\uparrow$ & FPS $\uparrow$ \\
\midrule
 64 & 98.3 & 98.1 & \underline{85.4} & 95.5 & \textbf{100} & 90.6 & \textbf{65.0} \\
 128 & 98.7 & \underline{98.2} & 85.1 & 96.2 & \textbf{100} & 91.2 & \underline{63.1} \\
\cellcolor{green!7} 256 & \cellcolor{green!7}\textbf{99.0} & \cellcolor{green!7}\underline{98.2} & \cellcolor{green!7}85.3 & \cellcolor{green!7}\textbf{96.8} & \cellcolor{green!7}\textbf{100} & \cellcolor{green!7}\underline{91.4} & \cellcolor{green!7}62.8 \\
 512 & \underline{98.9} & \textbf{98.3} & \textbf{85.6} & \underline{96.7} & \underline{99.9} & \textbf{91.5} & 59.5 \\
\bottomrule
\end{tabular}%
}
\end{table}

We investigate the impact of the candidate trajectory count $N$ on the overall driving performance. As reported in \cref{tab: app_traj_num}, we vary $N \in \{64, 128, 256, 512\}$ while maintaining all other hyperparameters constant.

Constraining $N$ to a minimal value of 64 yields a notable performance drop, since the limited prototype pool fails to encapsulate diverse multimodal maneuvers, thereby severely restricting the expressiveness of the action space. As $N$ increases, the driving metrics steadily improve but exhibit clear diminishing returns alongside a noticeable degradation in inference speed. Specifically, scaling $N$ from 256 to 512 provides a marginal 0.1\% performance gain at the expense of a 5.3\% speed reduction. Consequently, we adopt $N=256$ as our default configuration to secure the optimal balance between trajectory diversity and computational efficiency.

\subsection{Computational Efficiency Analysis}
\label{sec: app_efficiency}

\begin{wrapfigure}{r}{0.53\textwidth}
\vspace{-1.5em}
\includegraphics[width=0.53\textwidth]{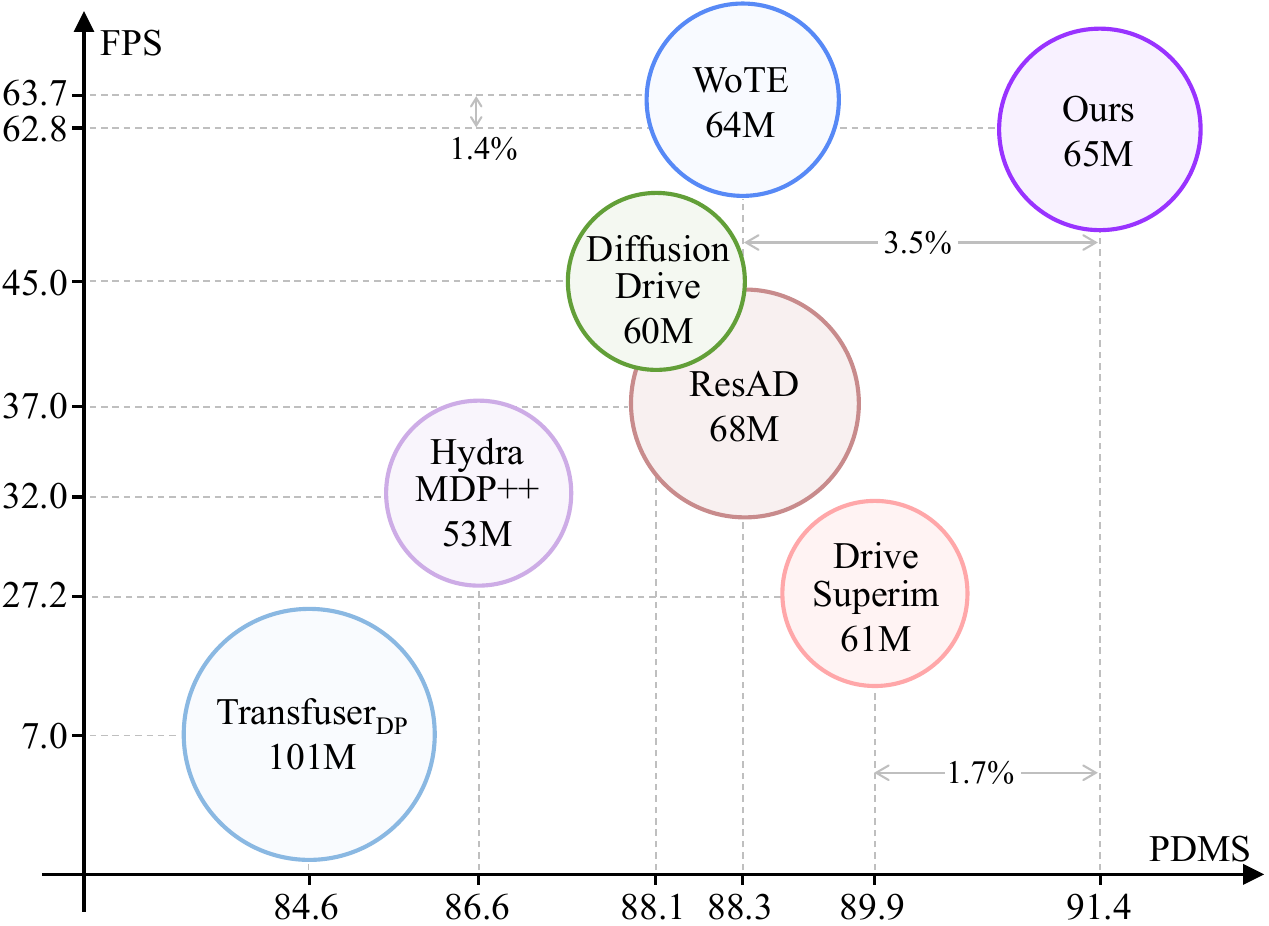}
\vspace{-1.5em}
\caption{Computational efficiency and performance comparisons.}
\label{fig: app_eff}
\vspace{-1.0em}
\end{wrapfigure}
In this section, we evaluate the computational efficiency of our framework against open-source methods and those reporting relevant metrics~\cite{prakash2021multi, li2025hydra++, yao2026drivesuprim, zheng2025resad, liao2025ddrive, li2025end}. 
As illustrated in \cref{fig: app_eff}, driven by lightweight cognitive distillation and an efficient world model design, our approach achieves the highest performance while maintaining a compact footprint and rapid inference, fully satisfying the rigorous constraints of real-time on-vehicle deployment. 
Specifically, compared to the fastest and most parameter-efficient baseline, WoTE~\cite{li2025end}, our method delivers a substantial $3.5\%$ performance improvement at a negligible cost, adding only $1.3\%$ in parameters and slowing down by a mere $1.4\%$. 
Furthermore, against the second-best performing method, DriveSuprim~\cite{yao2026drivesuprim}, our framework not only secures a $1.7\%$ performance margin but also achieves a staggering $130.9\%$ increase in inference speed.

\subsection{VLA Comparison and VLM Distillation Analysis}
\label{sec: app_vlm}

\begin{table}[h]
\centering
\caption{\small Comparison with VLA baselines and VLM distillation ablation on NAVSIM NAVTEST v1.}
\label{tab:ablvlm}
\resizebox{0.6\textwidth}{!}{
\begin{tabular}{l|ccccc|c}
\toprule
Method / VLM Source & NC $\uparrow$ & DAC $\uparrow$ & EP $\uparrow$ & TTC $\uparrow$ & C $\uparrow$ & PDMS $\uparrow$ \\
\midrule
\multicolumn{7}{c}{\cellcolor{gray!15}\textit{Comparison with Vision-Language-Action (VLA) Baselines}} \\
\midrule
AutoVLA~\cite{zhou2025autovla} & 98.4 & 97.1 & 81.9 & \textbf{97.1} & \underline{99.9} & 89.1 \\
ReCogDrive~\cite{li2025recogdrive} & 98.2 & 97.8 & 83.5 & 95.2 & 99.8 & 89.6 \\
DriveVLA-W0~\cite{li2025drivevla} & 98.7 & \textbf{99.1} & 83.3 & 95.3 & 99.3 & 90.2 \\
ExplorVLA~\cite{sheng2026explorevla} & \underline{98.8} & \underline{98.4} & 83.5 & 96.5 & \underline{99.9} & \underline{90.4} \\
\midrule
\multicolumn{7}{c}{\cellcolor{gray!15}\textit{Ablation on Different VLM Distillation Sources}} \\
\midrule
Qwen2.5-7B~\cite{bai2025qwen3} & 97.6 & 95.4 & 82.2 & 93.7 & \textbf{100} & 88.5 \\
Qwen3-8B~\cite{bai2025qwen3} & 97.8 & 95.7 & 82.0 & 93.9 & \textbf{100} & 88.3 \\
InternVL3-8B~\cite{zhu2025internvl3} & 97.6 & \underline{95.9} & \underline{83.7} & 94.4 & \underline{99.9} & 89.0 \\
\midrule
\cellcolor{green!7} Ours (Task-Aligned) & \cellcolor{green!7}\textbf{99.0} & \cellcolor{green!7}98.2 & \cellcolor{green!7}\textbf{85.3} & \cellcolor{green!7}\underline{96.8} & \cellcolor{green!7}\textbf{100} & \cellcolor{green!7}\textbf{91.4} \\
\bottomrule
\end{tabular}%
}
\end{table}

To comprehensively evaluate our approach, \cref{tab:ablvlm} extends the benchmarking to recent Vision-Language-Action (VLA) architectures~\cite{zhou2025autovla, li2025drivevla, li2025recogdrive, sheng2026explorevla} alongside different VLM distillation variants. 
To ensure equitable comparisons, all reported VLA baselines are strictly evaluated under a best-of-1 setting~\cite{zhou2025autovla}. 
While end-to-end VLAs exhibit strong semantic reasoning, our framework consistently surpasses them by grounding cognitive priors within explicit physical simulations. 
Furthermore, regarding the distillation source, the results reveal that generic, non-adapted VLMs yield only marginal performance variations. 
We therefore adopt InternVL3-8B~\cite{zhu2025internvl3} for its robust adaptability, explicitly fine-tuning it to inject domain-specific cognition into our policy. For rigorous benchmarking, all evaluated models undergo uniform LoRA fine-tuning (rank $r=16$, scaling factor $\alpha=32$, learning rate $2 \text{e-4}$). Crucially, this task-aligned adaptation unlocks substantial improvements, proving that domain-expert supervision fundamentally dictates the efficacy of cognitive distillation.

Beyond performance metrics, end-to-end VLAs are intrinsically bottlenecked by billion-scale parameter counts and seconds-level inference latency~\cite{zhou2025autovla, li2025drivevla}, precluding realistic edge deployment. Conversely, by completely decoupling the heavy cognitive engine during inference, our deployed policy operates on a compact million-scale parameter count. This structural elegance unlocks millisecond-level inference, bridging the critical gap between high-level cognitive reasoning and the stringent real-time execution demands of autonomous driving.

\subsection{Future BEV Map Prediction}
\label{sec: app_bev}

\begin{figure*}[!h]
\includegraphics[width=1.0\textwidth]{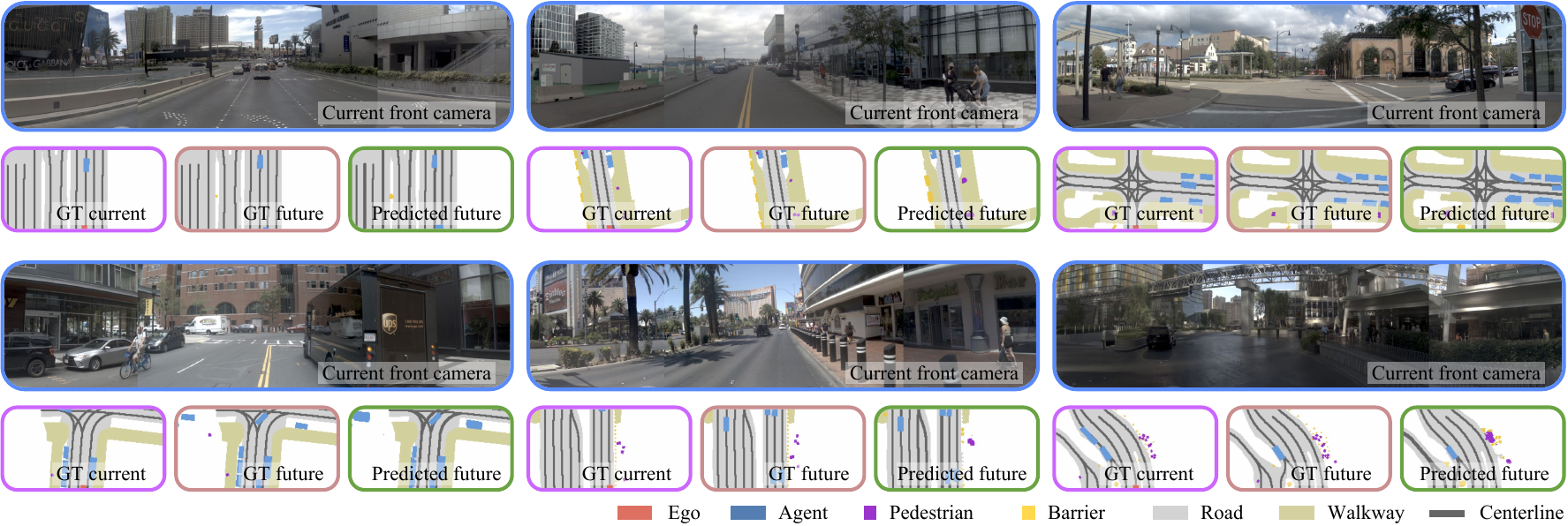}
\vspace{-1.5em}
\caption{Qualitative results of future BEV semantic map prediction.}
\label{fig: app_bev}
\end{figure*}

To visually validate our predictive capabilities, \cref{fig: app_bev} showcases the current front-view camera observation, the corresponding current BEV semantic map, and the predicted future BEV semantic map at the 4-second horizon. 
These visualizations corroborate the world model's proficiency in accurately forecasting the evolving spatial configurations of road boundaries, lane markings, and dynamic agents. 
While exceptionally small objects might present marginal boundary imprecision, their core semantic occupancies remain strictly preserved, ensuring that these localized artifacts never compromise the overarching structural integrity of the generated environments. Crucially, this predictive capability establishes a high-fidelity sandbox for the reinforcement learning phase. Grounded in this reliable spatial context, the dual-score trajectory evaluation ultimately empowers the policy to execute exceptionally safe and intention-aware planning decisions.

\subsection{Customizable Driving Behavior via Physical Reward Modulation}
\label{sec: person}

\begin{table}[h]
\centering
\caption{\small Effect of different weights of physical rewards on NAVSIM NAVTEST v1.}
\label{tab:ablrl}
\resizebox{0.67\textwidth}{!}{
\begin{tabular}{l|ccccc|c}
\toprule
Policy priorities & NC $\uparrow$ & DAC $\uparrow$ & EP $\uparrow$ & TTC $\uparrow$ & C $\uparrow$ & PDMS $\uparrow$ \\
\midrule
Speed and efficiency priority & 98.1 & 96.2 & 86.6 & 92.5 & 99.9 & 90.4 \\
Comfort and safety priority & 99.0 & 98.5 & 81.1 & 97.0 & 100 & 90.9 \\
Balanced priorities & 99.0 & 98.2 & 85.3 & 96.8 & 100 & 91.4 \\
\bottomrule
\end{tabular}%
}
\end{table}

In addition to high-level intent control via human language commands, our framework also supports microscopic preference customization directly through its physical reward mechanism.
In real-world autonomous driving deployments, humans exhibit highly diverse preferences. While some passengers prefer efficient and fast commutes, others prioritize smooth kinematics and conservative safety margins. Because our CoPhy framework explicitly models these factors through the physical reward formulation $\mathcal{R}_{Phy}$ during the reinforcement learning phase, it offers an elegant mechanism to accommodate these distinct preferences. By dynamically adjusting the weighting coefficients $\omega_{1...5}$, we can tailor the driving policy to execute specific driving priorities without requiring any structural modifications to the underlying model architecture.

Recall the physical reward formulation for the $i$-th candidate trajectory:
$$
    \mathcal{R}_{\texttt{Phy}}^{i} = \omega_{1} \mathcal{R}_{\texttt{NC}}^{i} + \omega_{2} \mathcal{R}_{\texttt{DAC}}^{i} + \omega_{3} \mathcal{R}_{\texttt{EP}}^{i} + \omega_{4} \mathcal{R}_{\texttt{TTC}}^{i} + \omega_{5} \mathcal{R}_{\texttt{Comf}}^{i}
$$

To ensure foundational driving safety, the weights for the hard boolean constraints, no collision ($\omega_{1}$) and drivable area compliance ($\omega_{2}$), remain strictly fixed across all variants. We adjust the remaining coefficients to define three distinct policy priorities.

\textit{(1) Balanced priority:} This represents our default configuration designed to strike an optimal balance between safety and efficiency. Following our primary experimental setup, the weights are set to $\omega_{3}=0.4$, $\omega_{4}=0.4$, and $\omega_{5}=0.2$.

\textit{(2) Comfort and safety priority:} This configuration prioritizes passenger comfort and extensive safety margins over task efficiency, making it highly suitable for elderly individuals or families with infants. By increasing the weight for Comfort, and decreasing the weight for Ego Progress (e.g., $\omega_{3}=0.1$, $\omega_{4}=0.4$, $\omega_{5}=1.0$), the policy is heavily incentivized to maintain larger follow distances and execute softer kinematic maneuvers.

\textit{(3) Speed and efficiency priority:} This configuration maximizes travel efficiency while maintaining baseline safety. By up-weighting Ego Progress and marginally relaxing the optimization for Time-to-Collision and Comfort (e.g., $\omega_{3}=1.0$, $\omega_{4}=0.2$, $\omega_{5}=0.1$), the policy actively seeks safe opportunities to overtake or maintain higher cruising speeds. This effectively addresses the inherent trade-off between safety and progress characteristic of standard imitation learning bounds.

As shown in \cref{tab:ablrl}, by simply modulating the reward coefficients, our framework produces distinctly different driving behaviors while consistently maintaining high overall performance, demonstrating the practical flexibility of CoPhy for diverse real-world deployment preferences.

\subsection{Limitations and Broader Impacts}
\label{sec: app_limit}
While our framework demonstrates robust performance, its real-world deployment must be rigorously contextualized within the systemic challenges inherent to the broader autonomous driving domain. 

First, a persistent limitation across current methodologies, including ours, is the inability to autonomously adapt to the highly divergent traffic regulations across global jurisdictions. This geographic generalization bottleneck inevitably necessitates redundant, region-specific policy fine-tuning. 

Furthermore, the rapid advancement of autonomous driving technologies significantly outpaces the evolution of corresponding legal and regulatory frameworks. This discrepancy creates critical ambiguities regarding accident liability and accountability in complex scenarios. 

Finally, the aggressive commercialization of these intelligent systems poses profound socioeconomic risks. The potential widespread displacement of professional human drivers could precipitate substantial job losses, inadvertently catalyzing wealth centralization and exacerbating societal inequity. Addressing these multidimensional challenges requires unprecedented collaborative efforts among technologists, policymakers, and ethicists to ensure the responsible integration of autonomous systems into global infrastructure.

\clearpage
\section*{NeurIPS Paper Checklist}

\begin{enumerate}

\item {\bf Claims}
    \item[] Question: Do the main claims made in the abstract and introduction accurately reflect the paper's contributions and scope?
    \item[] Answer: \answerYes{} 
    \item[] Justification: Please refer to Secs.~\ref{sec: method}, \ref{sec: exp}, and App.~\ref{sec: appen}.

\item {\bf Limitations}
    \item[] Question: Does the paper discuss the limitations of the work performed by the authors?
    \item[] Answer: \answerYes{} 
    \item[] Justification: Please refer to App.~\ref{sec: app_limit}.

\item {\bf Theory assumptions and proofs}
    \item[] Question: For each theoretical result, does the paper provide the full set of assumptions and a complete (and correct) proof?
    \item[] Answer: \answerYes{} 
    \item[] Justification: Please refer to Secs.~\ref{sec: method}, \ref{sec: exp}, and App.~\ref{sec: appen}.

    \item {\bf Experimental result reproducibility}
    \item[] Question: Does the paper fully disclose all the information needed to reproduce the main experimental results of the paper to the extent that it affects the main claims and/or conclusions of the paper (regardless of whether the code and data are provided or not)?
    \item[] Answer: \answerYes{} 
    \item[] Justification: Please refer to Secs.~\ref{sec: method}, \ref{sec: exp}, and App.~\ref{sec: appen}.

\item {\bf Open access to data and code}
    \item[] Question: Does the paper provide open access to the data and code, with sufficient instructions to faithfully reproduce the main experimental results, as described in supplemental material?
    \item[] Answer: \answerYes{} 
    \item[] Justification: Please refer to Secs.~\ref{sec: method}, \ref{sec: exp}, and App.~\ref{sec: appen}. The code will be made publicly available immediately after the paper is accepted and passes the commercial confidentiality review.

\item {\bf Experimental setting/details}
    \item[] Question: Does the paper specify all the training and test details (e.g., data splits, hyperparameters, how they were chosen, type of optimizer) necessary to understand the results?
    \item[] Answer: \answerYes{} 
    \item[] Justification: Please refer to Sec.~\ref{sec: exp} and App.~\ref{sec: appen}.

\item {\bf Experiment statistical significance}
    \item[] Question: Does the paper report error bars suitably and correctly defined or other appropriate information about the statistical significance of the experiments?
    \item[] Answer: \answerNA{} 
    \item[] Justification: Due to high computational costs and domain-specific paradigms, we have not provided error bars.

\item {\bf Experiments compute resources}
    \item[] Question: For each experiment, does the paper provide sufficient information on the computer resources (type of compute workers, memory, time of execution) needed to reproduce the experiments?
    \item[] Answer: \answerYes{} 
    \item[] Justification: Please refer to Sec.~\ref{sec: exp}.
    
\item {\bf Code of ethics}
    \item[] Question: Does the research conducted in the paper conform, in every respect, with the NeurIPS Code of Ethics \url{https://neurips.cc/public/EthicsGuidelines}?
    \item[] Answer: \answerYes{} 
    \item[] Justification: Our research conforms to the standards in every respect.

\item {\bf Broader impacts}
    \item[] Question: Does the paper discuss both potential positive societal impacts and negative societal impacts of the work performed?
    \item[] Answer: \answerYes{} 
    \item[] Justification: Please refer to App.~\ref{sec: app_limit}.
    
\item {\bf Safeguards}
    \item[] Question: Does the paper describe safeguards that have been put in place for responsible release of data or models that have a high risk for misuse (e.g., pre-trained language models, image generators, or scraped datasets)?
    \item[] Answer: \answerNA{} 
    \item[] Justification: Our model is a domain-specific driving policy and does not pose high misuse risks.

\item {\bf Licenses for existing assets}
    \item[] Question: Are the creators or original owners of assets (e.g., code, data, models), used in the paper, properly credited and are the license and terms of use explicitly mentioned and properly respected?
    \item[] Answer: \answerYes{} 
    \item[] Justification: We respect all assets, and have added citations wherever they are involved.

\item {\bf New assets}
    \item[] Question: Are new assets introduced in the paper well documented and is the documentation provided alongside the assets?
    \item[] Answer: \answerYes{} 
    \item[] Justification: Please refer to Sec.~\ref{sec: exp} and App.~\ref{sec: appen}. The code will be made publicly available immediately after the paper is accepted and passes the commercial confidentiality review.

\item {\bf Crowdsourcing and research with human subjects}
    \item[] Question: For crowdsourcing experiments and research with human subjects, does the paper include the full text of instructions given to participants and screenshots, if applicable, as well as details about compensation (if any)? 
    \item[] Answer: \answerNA{} 
    \item[] Justification: Our paper does not involve crowdsourcing nor research with human subjects.

\item {\bf Institutional review board (IRB) approvals or equivalent for research with human subjects}
    \item[] Question: Does the paper describe potential risks incurred by study participants, whether such risks were disclosed to the subjects, and whether Institutional Review Board (IRB) approvals (or an equivalent approval/review based on the requirements of your country or institution) were obtained?
    \item[] Answer: \answerNA{} 
    \item[] Justification: Our paper does not involve crowdsourcing nor research with human subjects.

\item {\bf Declaration of LLM usage}
    \item[] Question: Does the paper describe the usage of LLMs if it is an important, original, or non-standard component of the core methods in this research? Note that if the LLM is used only for writing, editing, or formatting purposes and does \emph{not} impact the core methodology, scientific rigor, or originality of the research, declaration is not required.
    \item[] Answer: \answerYes{} 
    \item[] Justification: Please refer to Secs.~\ref{sec: method}, \ref{sec: exp}, and App.~\ref{sec: appen}.

\end{enumerate}

\end{document}